%% file: AsyncDrop.tex
\documentclass[twoside]{article}

\usepackage[accepted]{aistats2023}
\usepackage{algorithm}
\usepackage{algorithmic}
\usepackage{graphicx}
\usepackage{booktabs}
\usepackage{enumitem}
\usepackage{tcolorbox}
\usepackage{wrapfig}
\usepackage{cuted}
\newcommand*{\inftybar}{\medskip \begin{center}\rule[.5ex]{15ex}{0.5pt} ~$\infty$~ \rule[.5ex]{15ex}{0.5pt}\end{center}}
\usepackage[pagebackref=true]{hyperref}
\renewcommand*{\backref}[1]{}
\renewcommand*{\backrefalt}[4]{[{\tiny%
    \ifcase #1 Not cited.%
          \or Cited on page~#2.%
          \else Cited on pages #2.%
    \fi%
    }]}
\definecolor{myblue2}{HTML}{4682B4}

\input{math_commands.tex}

\usepackage{hyperref}
\usepackage{url}

\usepackage[utf8]{inputenc} 
\usepackage[T1]{fontenc}    
\usepackage{hyperref}       
\usepackage{url}            
\usepackage{booktabs}       
\usepackage{amsfonts}       
\usepackage{nicefrac}       
\usepackage{microtype}      
\usepackage{xcolor}         

\begin{document}

%
\runningtitle{Efficient and Light-Weight Federated Learning via Asynchronous Distributed Dropout}

%

\twocolumn[

\aistatstitle{Efficient and Light-Weight Federated Learning \\ via Asynchronous Distributed Dropout}

\aistatsauthor{ Chen Dun \And Mirian Hipolito \And Chris Jermaine \And  Dimitrios Dimitriadis \And Anastasios Kyrillidis }

\aistatsaddress{ Rice University \And  Microsoft Research \And Rice University \And Microsoft Research \And Rice University } ]

\begin{abstract}
Asynchronous learning protocols have regained attention lately, especially in the Federated Learning (FL) setup, where slower clients can severely impede the learning process. Herein, we propose \texttt{AsyncDrop}, a novel asynchronous FL framework that utilizes dropout regularization to handle device heterogeneity in distributed settings. Overall, \texttt{AsyncDrop} achieves better performance compared to state of the art asynchronous methodologies, while resulting in less communication and training time overheads. The key idea revolves around creating ``submodels'' out of the global model, and distributing their training to workers, based on device heterogeneity. We rigorously justify that such an approach can be theoretically characterized. We implement our approach and compare it against other asynchronous baselines, both by design and by adapting existing synchronous FL algorithms to asynchronous scenarios. Empirically, \texttt{AsyncDrop} reduces the communication cost and training time, while matching or improving the final test accuracy in diverse non-i.i.d. FL scenarios.
\end{abstract}

\vspace{-0.2cm}
\section{Introduction}
\vspace{-0.2cm}

\textcolor{teal}{\textbf{Background on Federated Learning.}}
Federated Learning (FL) \cite{mcmahan2017communication, https://doi.org/10.48550/arxiv.1812.06127, https://doi.org/10.48550/arxiv.1910.06378} is a distributed learning protocol that has witnessed fast development the past demi-decade. 
FL deviates from the traditional distributed learning paradigms and 
allows the integration of edge devices ---such as smartphones \cite{stojkovic2022applied}, drones \cite{qu2021decentralized}, and IoT devices \cite{nguyen2021federated}--- in the learning procedure.
Yet, such real-life, edge devices are extremely \emph{heterogeneous} \cite{wang2021field}: they have drastically different specifications in terms of compute power, device memory and achievable communication bandwidths. 
Directly applying common \emph{synchronized} FL algorithms --such as FedAvg and FedProx 
\cite{https://doi.org/10.48550/arxiv.1812.06127,mcmahan2017communication} that require full model broadcasting and global synchronization-- results often in a ``stragglers'' effect \cite{nguyen2022federated,huba2022papaya, tandon2017gradient};
i.e., computationally powerful edge devices wait for slower ones during the synchronization step. 

\textcolor{teal}{\textbf{The ubiquitous synchronous training.}} One way to handle such issues is by utilizing \emph{asynchrony} instead of \emph{synchrony} in the learning process. 
To explain the main differences, let us first set up the background.
In a synchronous distributed algorithm, a global model is usually stored at a central server and is broadcast periodically to all the participating devices. 
Then, each device performs local training steps on its own model copy, before the device sends the updated model to the central server. 
Finally, the central server updates the global model by aggregating the received model copies. 
This protocol is followed in most FL algorithms, including the well-established FedAvg \cite{mcmahan2017communication}, FedProx \cite{https://doi.org/10.48550/arxiv.1812.06127}, FedNova \cite{wang2020tackling} and SCAFFOLD \cite{https://doi.org/10.48550/arxiv.1910.06378}.
The main criticism against synchronous learning could be that it often results in heavy communication/computation overheads and long idle/waiting times for workers. 

\textcolor{teal}{\textbf{Asynchrony and its challenges.}}
The deployment of a asynchronous learning method is often convoluted. 
In the past decade, HogWild! \cite{https://doi.org/10.48550/arxiv.1106.5730, liu2014asynchronous} has emerged as a general asynchronous distributed methodology, and has been applied initially in basic ML problems like sparse linear/logistic regression \cite{zhuang2013fast, yun2013nomad, hsieh2015passcode}.
Ideally, based on sparsity arguments, each edge device can independently update parts of the global model --that overlap only slightly with the updates of other workers-- in a lock-free fashion \cite{https://doi.org/10.48550/arxiv.1106.5730, liu2014asynchronous}.
This way, faster, more powerful edge workers do not suffer from idle waiting due to slower stragglers. 
Yet, the use of asynchrony has been a topic of dispute in distributed neural network training \cite{dean2012large, chen2016revisiting}. 
Asynchronous training often suffers from lower accuracy as compared to synchronized SGD, which results in the dominance of synchronized SGD in neural network training \cite{chen2016revisiting}. 

\textcolor{teal}{\textbf{Resurgence in asynchrony.}}
Recently, asynchronous methods have regained popularity, mainly due to the interest in applying asynchronous motions within FL on edge devices: 
the heterogeneity of edge networks, the ephemeral nature of the workers, the computational, communication and energy restriction of mobile devices are some impediments towards applying synchronous algorithms in realistic environments.
Yet, traditional off-the-shelf asynchronous distributed algorithms still have issues, which might be exacerbated in the FL setting.
As slower devices take longer local training time before updating the global model, this might result in inconsistent update schedules of the global model, compared to that of faster devices. 
This might have ramifications: $i)$ For FL on i.i.d. data, this will cause the gradient staleness problem and result in convergence rate decrease; and,
$ii)$ on non-i.i.d. data, this will result in a significant drop in global model final accuracy. 


As solutions, novel approaches on asynchronous FL propose weighted global aggregation techniques that take into consideration the heterogeneity of the devices \cite{https://doi.org/10.48550/arxiv.1903.03934, chen2019communication, 9407951}; yet, these methods often place a heavy computation/communication burden, as they rely on broadcasting full model updates to all the clients and/or the server. 
Other works monitor client speed to guide the training assignments \cite{li2021stragglers, chai2020fedat}.
Finally, recent efforts propose semi-asynchronous methods, where participating devices are selected and buffered in order to complete a semi-synchronous global update periodically \cite{huba2022papaya,wu2020safa}. 
A thorough discussion on the existing asynchronous methods in FL can be found in \cite{https://doi.org/10.48550/arxiv.2109.04269}.

\textcolor{teal}{\textbf{What is different in this work?}} 
As most algorithms stem from adapting asynchrony in synchronous FL, \textit{one still needs to broadcast the full model to all devices, following a data parallel distributed protocol \cite{farber1997parallel,raina2009large}, regardless of device heterogeneity}. This inspire us to ask a key question:

\begin{minipage}[t]{0.99\linewidth} 
\begin{tcolorbox}[colback=gray!5,colframe=green!40!black] 
\vspace{-0.2cm}
\textit{``Can we select submodels out of the global model and send these instead to each device, taking into account the device heterogeneity?''} \vspace{-0.2cm}
\end{tcolorbox} 
\end{minipage}  


We answer this question affirmatively, by proposing a novel distributed dropout method for FL.
We dub our method \texttt{AsyncDrop}.
Our approach assigns different submodels to each device\footnote{We consider both random assignment, as well as structured assignments, based on the computation power of the devices.}; empirically, such a strategy decreases the required time to converge to an accuracy level, while preserving favorable final accuracy.
This work attempts to reinstitute the discussion between synchrony and asynchrony in heterogeneous distributed scenarios, as in FL.
Our idea is based on the ideas of HogWild! \cite{https://doi.org/10.48550/arxiv.1106.5730, liu2014asynchronous} --in terms of sparse submodels-- and Independent Subnetwork Training (IST) \cite{yuan2022distributed, dun2022resist, liao2021convergence, wolfe2021gist} --where submodels are deliberately created for distribution, in order to decrease both computational and communication requirements. 

Yet, we deviate from these works: $i)$ The combination of HogWild! and IST ideas has not been stated and tested before this work, to the best of our knowledge. $ii)$ While HogWild!-line of work provides optimization guarantees, we consider the non-trivial, non-convex neural network setting and provide theoretical guarantees for convergence; such a result combines tools from asynchronous optimization \cite{https://doi.org/10.48550/arxiv.1106.5730, liu2014asynchronous}, Neural Tangent Kernel assumptions \cite{jacot2018neural}, dropout theory analysis \cite{liao2021convergence}, and focuses on convolutional neural networks \cite{lecun1989backpropagation}, deviating from fully-connected layer simplified scenarios. Finally, $iii)$ we provide system-level algorithmic solutions for our approach, mirroring best-practices found during our experiments.
Overall, the contributions of this work can be summarized as follows: \vspace{-0.3cm}
\begin{itemize}[leftmargin=*]
    \item We consider and propose \textit{asynchronous distributed dropout} (\texttt{AsyncDrop}) for efficient large-scale FL. Our focus is on non-trivial, non-convex ML models --as in neural network training-- and our framework provides specific engineering solutions for these cases in practice. \vspace{-0.15cm}
    \item We theoretically characterize and support our proposal with rigorous and non-trivial convergence rate guarantees. Currently, our theory assumes bounded delays; our future goal is to exploit recent developments that drop such assumptions \cite{koloskova2022sharper}. Yet, our theory already considers the harder case of neural network training, which is often omitted in existing theory results. \vspace{-0.15cm}
    \item We provide specific implementation instances and share ``best practices'' for faster distributed FL in practice. As a side-product, our preliminary results include baseline asynchronous implementations of many synchronous methods (such as FedAvg, FedProx, and more), that are not existent currently, to the best of our knowledge.
\end{itemize}

\vspace{-0.2cm}
\section{Problem Setup and Challenges}
\vspace{-0.2cm}
\textcolor{teal}{\textbf{Optimization in neural network training}}. 
We consider FL scenarios over \textit{supervised} neural network training: i.e., we optimize a loss function $\ell(\cdot, \cdot)$ over a dataset, such that the model maps unseen data to their true labels, unless otherwise stated.
For clarity, the loss $\ell(\mathbf{W}, \cdot)$ encodes both the loss metric and the neural architecture, with parameters $\mathbf{W}$.
Formally, given a data distribution $\mathcal{D}$ and $\left\{\mathbf{x}_i, y_i\right\} \sim \mathcal{D}$, where $\mathbf{x}_i$ is a data sample, and $y_i$ is its corresponding label, classical deep learning aims in finding $\mathbf{W}^\star$ as in: 
\vspace{-0.1cm}
\begin{align}
\mathbf{W}^\star &= \underset{\mathbf{W} \in \mathcal{H}}{\argmin} ~\left\{ \mathcal{L}(\mathbf{W}) := \tfrac{1}{n}\sum_{i = 1}^n \ell\left(\mathbf{W}, \left\{\mathbf{x}_i, y_i\right\} \right)\right\}, \nonumber 
\end{align}
where $\mathcal{H}$ denotes the model hypothesis class that ``molds'' the trainable parameters $\mathbf{W}$. 

The minimization above can be achieved by using different approaches, but almost all training is accomplished via a variation of \textit{stochastic gradient descent} (SGD) \cite{robbins1951stochastic}. 
SGD modifies the current guess $\mathbf{W}_t$ using stochastic directions $\nabla \ell_{i_t}(\mathbf{W}_i) := \nabla \ell(\mathbf{W}_i, \{\mathbf{x}_{i_t}, y_{i_t}\})$.
I.e., $\mathbf{W}_{t+1} \leftarrow \mathbf{W}_t - \eta \nabla \ell_{i_t}(\mathbf{W}_t).$
Here, $\eta > 0$ is the learning rate, and $i_t$ is a single or a mini-batch of examples. 
Most FL algorithms are based on these basic stochastic motions, like FedAvg \cite{mcmahan2017communication}, FedProx \cite{https://doi.org/10.48550/arxiv.1812.06127}, FedNova \cite{wang2020tackling} and SCAFFOLD \cite{https://doi.org/10.48550/arxiv.1910.06378}.

\textcolor{teal}{\textbf{FL formulation}}.
Let $S$ be the total number of clients in a distributed FL scenario.
Each client $i$ has its own local data $\mathcal{D}_i$ such that the whole dataset satisfies $\mathcal{D} = \cup_i \mathcal{D}_i$, and usually $\mathcal{D}_i \cap \mathcal{D}_j = \emptyset, \forall i \neq j$.
The goal of FL is to find a global model $\mathbf{W}$ that achieves good accuracy on all data $\mathcal{D}$, by minimizing the following optimization problem: \vspace{-0.15cm}
\begin{align*}
    \mathbf{W}^\star = \underset{\mathbf{W} \in \mathcal{H}}{\argmin} ~\left\{\mathcal{L}(\mathbf{W}) := \tfrac{1}{S}\sum_{i = 1}^S \ell\left(\mathbf{W}, \mathcal{D}_i\right)\right\}, \label{eq:fl_loss} \\[-15pt] \nonumber 
\end{align*}
where $\ell\left(\mathbf{W}, \mathcal{D}_i\right) = \tfrac{1}{|\mathcal{D}_i|} \sum_{\{\mathbf{x}_j, y_j\} \in \mathcal{D}_i} \ell\left(\mathbf{W}, \{\mathbf{x}_j, y_j\}\right)$.
With a slight abuse of notation, $\ell\left(\mathbf{W}, \mathcal{D}_i\right)$ denotes the \textit{local} loss function for user $i$, associated with a local model $\mathbf{W}_i$ (not indicated above), that gets aggregated with the models of other users. 
%
Herein, we consider both i.i.d. and non-i.i.d. cases, since local data distribution $\mathcal{D}_i$ can be heterogeneous and follow a non-i.i.d. distribution.

        \begin{algorithm}[!htp]
            \centering
            \caption{Meta Asynchronous FL}\label{alg:meta_async}
                \begin{algorithmic}
                    \STATE \textcolor{myblue2}{\textbf{Parameters}}: $T$ iters, $S$ clients, $l$ local iters., $\mathbf{W}$ as current global model, $\mathbf{W}_i$ as local model for $i$-th client, $\alpha \in (0, 1)$, $\eta$ step size.  \vspace{-0.5cm}
                    \inftybar
                    \STATE \vspace{-0.6cm}
                    \STATE $\mathbf{W}$ $\leftarrow$ randomly initialized global model.
                    
                    \STATE \textcolor{violet}{\begin{footnotesize}\texttt{//On each client $i$ asynchronously}:\end{footnotesize}}
                    \FOR{$t = 0, \dots, T-1$}
                    \STATE $\mathbf{W}_{i, t} \leftarrow \mathbf{W}$
                    \STATE \textcolor{violet}{\begin{footnotesize}\texttt{//Train $\mathbf{W}_i$ for $l$ iters. via SGD}\end{footnotesize}}
                        \FOR{$j = 1, \dots, l$}
                        \STATE $\mathbf{W}_{i, t} \leftarrow \mathbf{W}_{i, t} - \eta \frac{\partial \mathcal{L}}{\overline{\mathbf{W}_{i, t}}}$
                        \ENDFOR
                    \STATE \textcolor{violet}{\begin{footnotesize}\texttt{//Write local to global model}\end{footnotesize}}
                    \STATE $\mathbf{W} \leftarrow (1-\alpha) \cdot \mathbf{W}+\alpha \cdot \mathbf{W}_{i, t}$
                    \ENDFOR        
            \end{algorithmic}
        \end{algorithm}
        \vspace{-0.3cm}
        
\textcolor{teal}{\textbf{Details of asynchronous training.}}
An abstract description of how asynchronous FL operates is provided in Algorithm \ref{alg:meta_async}. 
In particular, given a number of server iterations $T$, each client $i$ gets the updated global model $\mathbf{W}_t$ from the server, and further locally trains it using $\mathcal{D}_i$ for a number of local iterations $l$.\footnote{Details on the use of the optimizer, how it is tuned with respect to step size, mini-batch size, etc. are intentionally hidden at this point of the discussion.} 
Asynchronous FL assumes each client has different computation power and communication bandwidth; this can be abstracted by different wall-clock times required to finish a number of local training iterations. 
Thus, when client $i$ has completed its round, the updated model is shared with the server to be aggregated, before the next round of communication and computation starts for client $i$.
\textit{This is different from classical synchronous FL, where the global model is updated only when all participating clients finish (or time-out) certain local training iterations. }

\vspace{-0.2cm}
\section{Challenges in Asynchronous FL and Related Work} \label{Sec:Challanges}
\vspace{-0.2cm}
\textcolor{teal}{\textbf{Challenges.}}
Asynchronous steps often lead to inconsistent update schedules of the global model and are characterized by gradient staleness and drifting. 
Real-life FL applications include edge devices with limited communication and computation capabilities (e.g., how often and fast they can connect with the central server, and how powerful as devices they are).
For instance, edge devices such as IoT devices or mobile phones \cite{nguyen2021federated}, might only be able to communicate with the server within short time windows, due to network conditions or user permission policy. 

\begin{figure}[H]
\vspace{-0.3cm}
\centering
\includegraphics[width=1\linewidth]{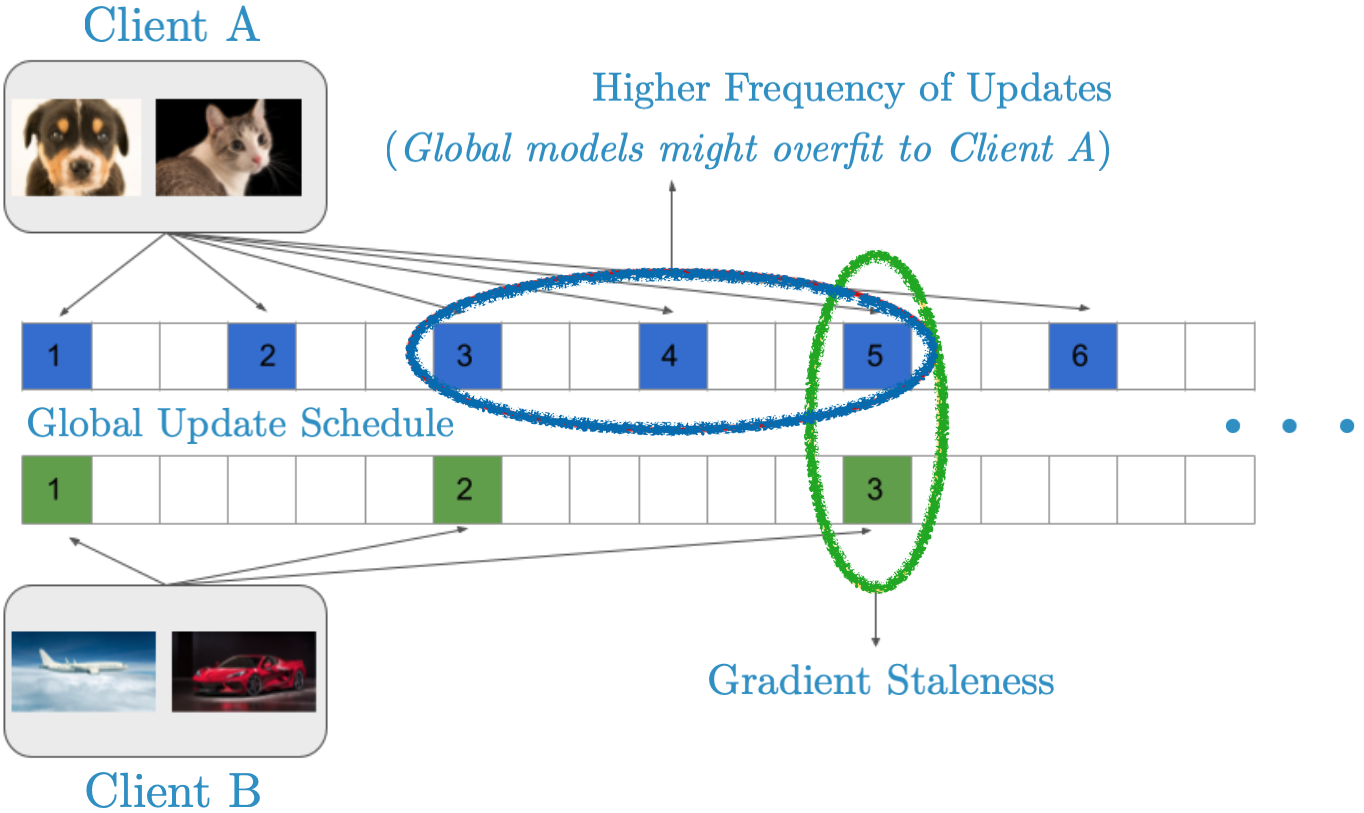}
\vspace{-0.5cm}
\caption{Potential issues in asynchronous FL.}
\vspace{-0.5cm}
\label{fig:async_overview}
\end{figure}
Consider the toy setting in Figure \ref{fig:async_overview}.
The two clients (Clients A and B) have a significantly different update schedule on the global model: Here, Client A has higher computational power or communication bandwidth --compared to client B-- potentially leading to model drifting, lack of fair training and more severe gradient staleness. 
On top, consider these two clients having different local (non-i.i.d.) data distributions. 


\textcolor{teal}{\textbf{Related Work.}}
The issue of model drifting due to data ``non-iidness'' is a central piece in FL research. 
Algorithms, such as FedProx \cite{https://doi.org/10.48550/arxiv.1812.06127}, utilize regularization to constrain local parameters ``drifting" away from previous global models. 

The gradient staleness problem has been widely studied in asynchronous FL, like in \cite{https://doi.org/10.48550/arxiv.1903.03934, chen2019communication, 9407951, li2021stragglers, chai2020fedat}. 
These approaches can be summarized as weighted asynchronous FedAvg protocols, in which the weight of each local client update is proportional to the ``capabilities'' of the client. 
This should decrease the negative impact from stale gradients by slower clients. 
Semi-asynchronous methods have been proposed \cite{huba2022papaya, wu2020safa}; yet, they require fast clients to wait until all other clients' updates are completed, in order to receive the updated model for the next round of local training.

Finally, numerous quantization \cite{commeff-sgd, double-quant} and sparsification \cite{sparse-comm, linear-speed-quant} techniques have been proposed for reducing computation and communication costs in FL. 

\vspace{-0.2cm}
\section{Asynchronous Distributed Dropout}
\vspace{-0.2cm}
\textcolor{teal}{\textbf{(Distributed) Dropout}}.
Dropout \cite{wan2013regularization, srivastava2014dropout, gal2016dropout, courbariaux2015binaryconnect} is a widely-accepted regularization technique in deep learning. 
The procedure of Dropout is as follows: per training round, a random mask over the parameters is generated; this mask is used to nullify part of the neurons in the neural network for this particular iteration. 
Variants of dropout include the drop-connect \cite{wan2013regularization}, multisample dropout \cite{inoue2019multi}, Gaussian dropout \cite{wang2013fast}, and the variational dropout \cite{kingma2015variational}. 

The idea of dropout has also been used in efficient distributed and/or FL scenarios.
\cite{horvath2021fjord} introduces FjORD and the Ordered Dropout, a \textit{synchronous} distributed dropout technique that leads to ordered, nested representation of knowledge in models, and enables the extraction of lower footprint submodels without the need of retraining. 
Such submodels are more suitable in client heterogeneity, as they adapt submodel's width to the client’s capabilities.
See also Nested Dropout \cite{rippel2014learning} and HeteroFL \cite{diao2020heterofl}.

\vspace{-0.2cm}
\begin{algorithm}[!h]
            \centering
            \caption{Asynchronous dropout (\texttt{AsyncDrop})} \label{alg:drop_async}
                \begin{algorithmic}
                    \STATE \textcolor{myblue2}{\textbf{Parameters}}: $T$ iters, $S$ clients, $l$ local iters., $\mathbf{W}$ as current global model, $\mathbf{W}_i$ as local model for $i$-th client, $\alpha \in (0, 1)$, $\eta$ step size.  \vspace{-0.5cm}
                    \inftybar 
                    \STATE \vspace{-0.6cm}
                    \STATE $\mathbf{W}$ $\leftarrow$ randomly initialized global model.
                    
                    \STATE \textcolor{violet}{\begin{footnotesize}\texttt{//On each client $i$ asynchronously}:\end{footnotesize}}
                    \FOR{$t = 0, \dots, T-1$}
                    \STATE \textcolor{teal}{Generate mask $\mathbf{M}_{i, t}$}
                    \STATE $\mathbf{W}_{i, t} \leftarrow \mathbf{W}_t \textcolor{teal}{~\odot~ \mathbf{M}_{i, t}}$
                    \STATE \textcolor{violet}{\begin{footnotesize}\texttt{//Train $\mathbf{W}_{i,t}$ for $l$ iters. via SGD}\end{footnotesize}}
                        \FOR{$j = 1, \dots, l$}
                        \STATE $\mathbf{W}_{i, t} \leftarrow \mathbf{W}_{i, t} - \eta \frac{\partial \mathcal{L}}{\overline{\mathbf{W}_{i, t}}}$
                        \ENDFOR
                    \STATE \textcolor{violet}{\begin{footnotesize}\texttt{//Write local to global model}\end{footnotesize}}
                    \STATE $\mathbf{W}_{t+1} \leftarrow \mathbf{W}_t \textcolor{teal}{~\odot~ (\mathbf{M}_{i, t})^c}$
                    \STATE \quad \quad \quad \quad \quad \quad $ + ((1-\alpha) \cdot \mathbf{W}_t+\alpha \cdot \mathbf{W}_{i, t}) \textcolor{teal}{~\odot~ \mathbf{M}_{i, t}}$
                    \ENDFOR        
            \end{algorithmic}
        \end{algorithm}

\begin{figure*}[!ht]
\vspace{0.1cm}
\centering
\includegraphics[width=0.85\linewidth]{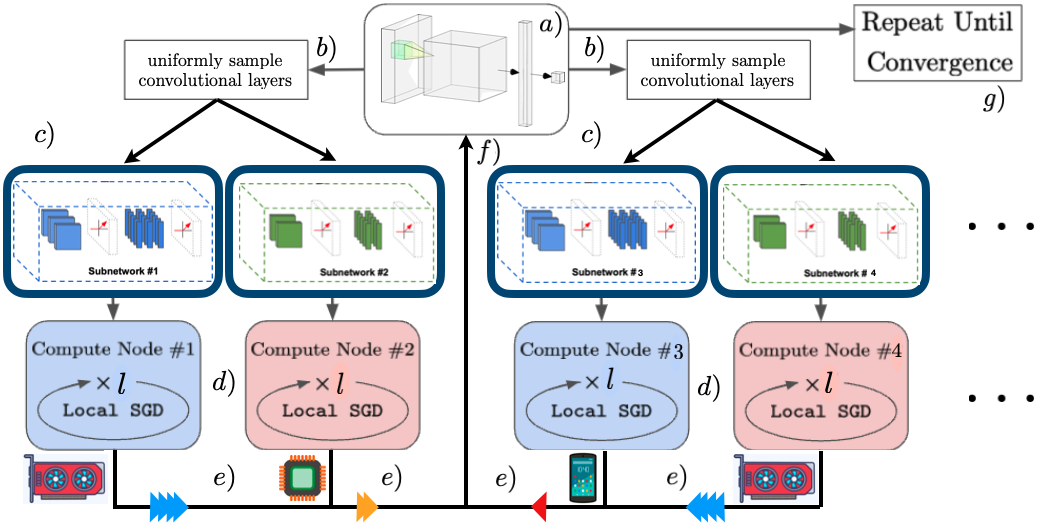}
\vspace{-0.5cm}
\caption{Schematic representation of \texttt{AsyncDropout}. $a)$ This is a simple representation of a CNN model. Our algorithm applies for arbitrary depth of CNNs (ResNets) as well as other architectures (MLPs, LSTMs, etc); here we restrict to a shallow CNN for illustration purposes. $b)$ Per request, random sub-sampled CNN models are created that result into different \textit{subnetworks}. $c)$ These submodels are distributed to devices with different computational capabilities (here GPU, CPU, or a smartphone). $d)$ Without loss of generality, we assume that all devices train locally the submodel for $l$ iterations. $e)$ However, each device finishes local training in different timestamps (shown as different colored arrows: \textcolor{red}{red}: slow speed; \textcolor{orange}{orange}: moderate speed; \textcolor{blue}{blue}: fast speed). $f)$ Yet, the global model is asynchronously updated and new submodels are created without global synchronization. $g)$ The above procedure is repeated till convergence.}
\label{fig:async_dropout_scheme}
\vspace{-0.4cm}
\end{figure*}

\textcolor{teal}{\textbf{Our proposal and main hypothesis}}.
We focus on the \emph{asynchronous version of distributed dropout.}
We study theoretically whether asynchrony provably works in non-trivial non-convex scenarios --\textit{as in training neural networks}-- with random masks that generate submodels for each worker.
The algorithm is described in Algorithm \ref{alg:drop_async}, dubbed as \texttt{AsyncDrop}, and is based on recent distributed protocols \cite{yuan2022distributed, dun2022resist, liao2021convergence, wolfe2021gist}; key features are highlighted in \textcolor{teal}{teal-colored text}.
The main difference from Algorithm \ref{alg:meta_async} is that Algorithm \ref{alg:drop_async} splits the model vertically per iteration,
where each submodel contains all layers of the neural network, but only with a (non-overlapping) subset of neurons being active in each layer. Multiple local SGD steps can be performed without the need for the workers to communicate. 
See also Figure \ref{fig:async_dropout_scheme} for a schematic representation of asynchronous distributed dropout for training a CNN.

\textcolor{teal}{\textbf{Theoretical Results}}.
We are interested in understanding whether such a combination of asynchronous computing and dropout techniques lead to convergence and favorable results: \textit{given the variance introduced by both asynchronous updates and training of submodels, it is not obvious whether --and under which conditions-- such a protocol would work.}

For ease of presentation and clarity of results, we analyse a one-hidden-layer CNN, and show convergence with random filter dropout.
Consider a training dataset $(\X,\y) = \{(\x_i, y_i)\}_{i=1}^n$, where each $\x_i\in\R^{\hat{d}\times p}$ is an image and $y_i$ being its label. 
Here, $\hat{d}$ is the number of input channels and $p$ the number of pixels. 
Let $q$ denote the size of the filter, and let $m$ be the number of filters in the first layer.
Based on previous work \cite{du2019gradient}, we let $\hat{\phi}(\cdot)$ denote the patching operator with $\hat{\phi}(x) \in\R^{q\hat{d}\times p}$. Consider the first layer weight $\W\in\R^{m\times q\hat{d}}$, and second layer (aggregation) weight  $\mathbf{a}\in\R^{m\times p}$. 
We assume that only the first layer weights $\W$ is trainable. 
The CNN trained on the means squared error has the form: \vspace{-0.3cm}

\begin{small}\begin{align*}
    f(\mathbf{x},\mathbf{W}) = \inner{\mathbf{a},{\sigma\paren{\W\hat{\phi}\paren{\x}}}};, \mathcal{L}\paren{\W} = \norm{f(\X,\W) - \y}_2^2,
\end{align*}
\end{small}

where $f(\mathbf{x}, \cdot)$ denotes the output of the one-layer CNN for input $\mathbf{x}$, and $\mathcal{L}(\cdot)$ is the loss function.
We use the $\ell_2$-norm loss for simplicity.
We make the following assumption on the training data and the CNN weight initialization.
\begin{asmp}
(Training Data)
\label{data_asump}
Assume that for all $i\in[n]$, we have $\norm{\x_i}_F = q^{-\frac{1}{2}}$ and $|y_i|\leq C$ for some constant $C$. Moreover, for all $i,i'\in[n]$ we have $\x_i\neq \x_{i'}$. 
\end{asmp}

Note that this can be satisfied by normalizing the data. For simplicity of the analysis, let $d: = q\hat{d}$.

\begin{asmp}
(Initialization)
\label{init_asump}$\w_{0,i}\sim\mathcal{N}\paren{0, \kappa^2\mathbf{I}}$ and $a_{i, i'}\sim\left\{\pm\frac{1}{p\sqrt{m}}\right\}$ for $i\in[m]$ and $i'\in[p]$.
\end{asmp}

In an asynchronous scenario, the neural network weight is updated with stale gradients due to the asynchronous updates, where $\delta_t$ is the delay at training step $t$. \textit{We assume $\delta_t$ is bounded by a constant $E$.}
Then, a simple version of gradient descent under these assumptions looks like:
\begin{align*}
    \W_t=\W_t-\eta \nabla_{\W}\mathcal{L}\paren{\W_{t-\delta_t}}, \quad \delta_t \leq E,
\end{align*}
where $\W_{t-\delta_t}$ indicates that the gradient is evaluated on a earlier version of the model parameters. 
Given the above, we provide the following guarantees: \vspace{-0.2cm}
\begin{thm}
\textit{Let $f(\cdot, \cdot)$ be a one-hidden-layer CNN with the second layer weight fixed. 
Let $\mathbf{u}_t$ abstractly represent the output of the model after $t$ iterations, over the random selection of the masks.
Let $E$ denotes the maximum gradient delay/staleness.
Let $\xi$ denote the dropout rate ($\xi = 1$ dictates that all neurons are selected), and denote $\theta = 1 - (1 - \xi)^S$ the probability that a neuron is active in at least one subnetwork.
Assume the number of hidden neurons satisfies
$m = \Omega\paren{\max\{\frac{n^4K^2}{\lambda_0^4\delta^2}\max\{n, d\},\frac{n}{\lambda_0}\}}$ and the step size satisfies $\eta = O\paren{\frac{\lambda_0}{n^2}}$.
Let $\kappa$ be a proper initialization scaling factor, and it is considered constant.
We use $\lambda_0$ to denote the smallest eigenvalue of the Neural Tangent Kernel matrix.
Let Assumptions 1 and 2 be satisfied. Then, the following convergence rate guarantee is proved to be satisfied:}
\end{thm}

\begin{strip}
\hrule
\begin{align*}
\E_{\mathbf{M}_t}\left[\norm{\U_{t+1} - \y}_2^2\right] & \leq \textcolor{red}{\paren{1 - \frac{\theta\eta\lambda_0}{4}}^{t}\norm{\U_0- \y}_2^2} \\
    & \quad\quad\quad + O\left(\textcolor{violet}{\frac{\theta\eta\lambda_0^3 \xi^2 \kappa^2 E^2}{n^2}} + \textcolor{orange}{\frac{\xi^2(1-\xi)^2\theta\eta n^3\kappa^2d}{m\lambda_0}} + \textcolor{violet}{\frac{\eta^2\theta^2 n \kappa ^2 \lambda_0 \xi^4 E^2}{m^{4}}} + \textcolor{orange}{\frac{\xi^2(1-\xi)^2\theta^2\eta^2 n^2  \kappa^2 d}{m^3\lambda_0 }} \right. \\
    &\left. \quad\quad\quad\quad\quad + \textcolor{violet}{\frac{\xi^2(1-\xi)^2\theta^2\eta^2 \kappa^2 \lambda_0 E^2}{m^3}} + \textcolor{orange}{\frac{\xi^2(1-\xi)^2\theta^2\eta^2 n^2  \kappa^2 d}{m^2\lambda_0}}+\textcolor{orange}{\frac{n\kappa^2\paren{\theta-\xi^2}}{S}}\right) \\
\end{align*}
\vspace{-0.8cm}
\hrule
\end{strip}

\begin{table*}[!htp]
\centering
\caption{Test accuracy of asynchronous FL baselines vs. \texttt{AsyncDrop} on non-i.i.d. CIFAR100 data over $>100$ clients. We also report the time and communication overhead to reach a certain target accuracy: 
``Time for \texttt{XX}\% Accuracy'' denotes the second lowest test accuracy among all baselines as the target accuracy.} 
\begin{footnotesize}
\begin{tabular}{ccccc}
\toprule
    &   Max. Test Accuracy & Time for 32\% Accuracy & Time Overhead & Comm. Overhead  \\ \midrule
    Sync. FedAvg & 60.47  & 3890.6s & +69.38\% & +42.2\% \\  \midrule
    Async. FedAvg & 32.47 $\pm$ 1.89 & 3062.5s & +33.00\% & +15.56\% \\
    Async. Fed-Weighted-Avg & 32.98 $\pm$ 1.71 & 3062.5s & +33.00\% & +15.56\%\\
    Async. FedProx & 35.75 $\pm$ 0.61 & 3062.5s & +33.33\% & +15.56\% \\
    Async. FjORD & 12.07 $\pm$ 0.83 & N/A & N/A & N/A \\ \midrule
    \texttt{AsyncDrop} & \textcolor{teal}{\textbf{35.93}} $\pm$ 0.92  & \textcolor{teal}{\textbf{2296.8s}} & \textcolor{teal}{\textbf{[Best]}} & \textcolor{teal}{\textbf{[Best]}} \\
 \bottomrule
\end{tabular}
\end{footnotesize}
\vspace{-0.3cm}
\label{cifar100_noniid_results_vanilla}
\end{table*}

\textcolor{teal}{\textbf{Remark \#1.}}
This theorem states that the sum of the expected weight differences in the $t$-th iteration (i.e., $\E_{\mathbf{M}_t}[\norm{\U_{t+1} - \y}_2^2]$) converges linearly to zero, as dictated by the red term --$\textcolor{red}{\paren{1 - \frac{\theta\eta\lambda_0}{4}}^{t}\norm{\U_0- \y}_2^2}$-- up to an error neighborhood, denoted with the Big-Oh notation term on the right hand side of the expression.
Focusing on the latter, there are two types of additive errors: $i)$ the \textcolor{orange}{orange-colored} terms origin from the dropout analysis: the term $1 - \xi$ is often called as ``dropout rate'' (when $\xi = 0$, no neurons are selected and the loss hardly
decreases, while when $\xi = 1$, all neurons are selected, and the \textcolor{orange}{orange-colored} terms disappear). 
$ii)$ the \textcolor{violet}{violet-colored terms} origin from the asynchronous analysis: when $E = 0$ (i.e., we boil down to synchronous computations), these terms also disappear). 

\textcolor{teal}{\textbf{Remark \#2.}}
Beyond the above extreme cases, we observe that the error region terms can be controlled by algorithmic and model-design choices: e.g., when the size of the dataset $n$ increases, the first term $\textcolor{violet}{\frac{\theta\eta\lambda_0^3 \xi^2 \kappa^2 E^2}{n^2}}$ can be controlled; for sufficiently wide neural network, the terms with $m$ in the denominator can be made arbitrarily small; finally, notice that increasing the number of subnetworks $S$ will drive the last term in the bound zero.

\textcolor{teal}{\textbf{Vanilla asynchronous distributed dropout in practice}}. 
We test vanilla \texttt{AsyncDrop} with 25\% Dropout rate in a FL setting with 104 heterogeneous clients and based on non-i.i.d. CIFAR100 dataset distribution. 
Beyond the extensions of FL baselines to the asynchronous setting (FedAvg, Fed-Weighted-Avg and FedProx), we further extend the work in \cite{horvath2021fjord} into Async. FjORD; further details in Sec \ref{Sec:Experiments}.  
As shown in Table \ref{cifar100_noniid_results_vanilla}, \texttt{AsyncDrop} indeed shows improvements in three components: \textit{final global model test accuracy, training time and communication cost}. 
These preliminary results demonstrate that \texttt{AsyncDrop} is on track to improve upon global model drifting, gradient staleness and computation/communication cost; all within the same single method.

\vspace{-0.2cm}
\section{Smart Partition in \texttt{AsyncDrop} for FL}
\vspace{-0.2cm}


\textcolor{teal}{\textbf{Going beyond \texttt{AsyncDrop}}}.
Despite theoretical support, \texttt{AsyncDrop} does not count device heterogeneity when submodels are assigned to the various workers. 
This could result into slower convergence rates and/or lower final model accuracy compared to other asynchronous methods \cite{https://doi.org/10.48550/arxiv.1903.03934, chen2019communication, 9407951, li2021stragglers, chai2020fedat} that carefully handle such cases. 
With non-i.i.d. data, more frequent updates by faster devices could lead to model drifting on local data. 
\emph{These facts suggest a more careful handling of model splitting and model distribution among heterogeneous workers.}

        \begin{algorithm}[!ht]
            \centering
            \caption{\texttt{Hetero \!\!\!\!\! AsyncDrop} for Asynchronous FL}\label{alg:async_IST_smart}
                \begin{algorithmic}
                \STATE \textcolor{myblue2}{\textbf{Parameters}}: $T$ iters, $S$ clients, $l$ local iters., $\mathbf{W}$ as current global model, $\mathbf{W}_i$ as local model for $i$-th client, \textcolor{teal}{$\eta_{g}$ as global LR, \textcolor{teal}{$v(\cdot)$ weight score function,} $\psi(i)$ computes the computation capacity of $i$-th worker, $\varphi(\mathbf{W}, \psi(i), v(\cdot))$ is the \texttt{Smart \!\!\!\!\! Dropout} function that creates the mask, based on worker capacity $\psi$ and score $v$}, $\alpha \in (0, 1)$. \vspace{-0.4cm}
                    \inftybar
                    \STATE \vspace{-0.5cm}
                    \STATE $\mathbf{W}$ $\leftarrow$ randomly initialized global model.
                    
                    \STATE \textcolor{violet}{\begin{footnotesize}\texttt{//On each client $i$ asynchronously}:\end{footnotesize}}
                    \FOR{$t = 0, \dots, T-1$}
                    
                    \STATE \textcolor{violet}{\begin{footnotesize}\texttt{//For $i$-th fastest worker, $\varphi(\cdot)$ drops weights with $i$-th largest $v(\cdot)$ score} \end{footnotesize}}
                    \STATE \textcolor{teal}{Generate mask $\mathbf{M}_{i, t} = \varphi(\mathbf{W}_t, \psi(i), v(\cdot))$}
                    \STATE $\mathbf{W}_{i, t} \leftarrow \mathbf{W}_t \textcolor{teal}{~\odot~ \mathbf{M}_{i, t}}$
                    \STATE \textcolor{violet}{\begin{footnotesize}\texttt{//Train $\mathbf{W}_{i,t}$ for $l$ iters. via SGD}\end{footnotesize}}
                        \FOR{$j = 1, \dots, l$}
                        \STATE $\mathbf{W}_{i, t} \leftarrow \mathbf{W}_{i, t} - \eta_g \frac{\partial \mathcal{L}}{\overline{\mathbf{W}_{i, t}}}$
                        \ENDFOR
                    \IF{$i$-th client is fastest}
                        \STATE \textcolor{teal}{Update $\eta_g$}
                    \ENDIF
                    \STATE \textcolor{violet}{\begin{footnotesize}\texttt{//Write local to global model}\end{footnotesize}}
                    \STATE $\mathbf{W}_{t+1} \leftarrow \mathbf{W}_t \textcolor{teal}{~\odot~ (\mathbf{M}_{i, t})^c}$
                    \STATE \quad \quad \quad \quad \quad \quad $ + ((1-\alpha) \cdot \mathbf{W}_t+\alpha \cdot \mathbf{W}_{i, t}) \textcolor{teal}{~\odot~ \mathbf{M}_{i, t}}$
                    \STATE \textcolor{violet}{\begin{footnotesize}\texttt{//Update score $q$}\end{footnotesize}}
                    \STATE $v(\mathbf{W}_{t+1}^{j}) = \norm{\mathbf{W}_{t+1}^j-\mathbf{W}_{0}^j}_1$, $\forall~j \in \mathcal{J}$
                    \ENDFOR  
            \end{algorithmic}
        \end{algorithm}

\textcolor{teal}{\textbf{\texttt{Hetero \!\!\!\!\! AsyncDrop}: An improved asynchronous solution}}.
We propose to ``balance'' the contribution from different devices with the \texttt{Hetero \!\!\!\!\! AsyncDrop} method.
Briefly, \texttt{Hetero \!\!\!\!\! AsyncDrop} assigns different weights to different devices, based on the rate update of the weights, and the computation power of the devices. 
The description of \texttt{Hetero \!\!\!\!\! AsyncDrop} is provided in Algorithm \ref{alg:async_IST_smart}.
We assign model weights that are less updated to the faster devices. 
Similarly, we assign model weights that are updated more often -- and, thus, converging faster-- to the slower devices. 
The premise behind such a protocol is that \emph{all weights, eventually, will be updated/will converge with a similar rate.}


\begin{table*}[!t]
\centering
\caption{Test accuracy of asynchronous FL baselines vs. (\texttt{Hetero}) \texttt{AsyncDrop} using a ResNet architecture on non-i.i.d CIFAR10, CIFAR100 and FMNIST data over $>100$ clients. We report the time and communication overhead to reach a certain target accuracy: 
``Time for \texttt{XX}\% Accuracy'' denotes the second lowest test accuracy among all baselines as the target accuracy. \textcolor{teal}{Teal colored text} indicates favorable performance; \textcolor{red}{red colored text} indicates high variance in performance.} 
\begin{footnotesize}
\begin{tabular}{ccccc}
\toprule
    \texttt{CIFAR10} &   Max. Test Accuracy & Time for 35\% Accuracy & Time Overhead & Comm. Overhead  \\ \midrule
    Async. FedAvg & 45.79 $\pm$ \textcolor{red}{\textbf{7.9}} & 2105.5s & +33.34\% & +15.56\% \\
    Async. Fed-Weighted-Avg & 46.51 $\pm$ \textcolor{red}{\textbf{6.8}} & 2105.5s & +33.34\% & +15.56\% \\
    Async. FedProx & 43.97 $\pm$ 1.35 & 2296.9s & +45.46\% & +26.06\% \\
    Async. FjORD & 23.14 $\pm$ 0.90 & N/A & N/A & N/A \\
    FedBuff & 35.81 $\pm$ \textcolor{red}{\textbf{11.83}} & 3012.9s & +89.98\% & +35.75\% \\ \midrule
    \texttt{Hetero AsyncDrop} & \textcolor{teal}{\textbf{50.67}} $\pm$ 1.75  & \textcolor{teal}{\textbf{1579.1s}} & \textcolor{teal}{\textbf{[Best]}}  & \textcolor{teal}{\textbf{[Best]}} \\ 
    \texttt{AsyncDrop} & 48.98 $\pm$ \textcolor{red}{\textbf{3.87}} & 2009.7s & +27.27 \% & +27.27\%
    \\
 \bottomrule
 \bottomrule
   \texttt{CIFAR100} &   Max. Test Accuracy & Time for 32\% Accuracy & Time Overhead & Comm. Overhead  \\ \midrule
    Async. FedAvg & 32.47 $\pm$ 1.89 & 3062.5s & +33.38\% & +15.56\% \\
    Async. Fed-Weighted-Avg & 32.98 $\pm$ 1.71 & 3062.5s & +33.38\% & +15.56\%\\
    Async. FedProx & 35.75 $\pm$ 0.61 & 3062.5s & +33.38\% & +15.56\%\\
    Async. FjORD & 12.07 $\pm$ 0.83 & N/A & N/A & N/A \\
    FedBuff & \textcolor{teal}{\textbf{41.91}} $\pm$ \textcolor{red}{\textbf{3.80}} & 4250.1s & +85.03\% & +42.22\%\\ 
    \midrule
    \texttt{Hetero AsyncDrop} & 37.26 $\pm$ 0.93  & \textcolor{teal}{\textbf{2296.8s}} & \textcolor{teal}{\textbf{[Best]}}  & \textcolor{teal}{\textbf{[Best]}} \\
    \texttt{AsyncDrop} & 35.93 $\pm$ 0.93  & \textcolor{teal}{\textbf{2296.8s}} & \textcolor{teal}{\textbf{[Best]}}  & \textcolor{teal}{\textbf{[Best]}} \\
 \bottomrule
 \bottomrule
    \texttt{FMNIST} &   Max. Test Accuracy & Time for 57\% Accuracy & Time Overhead & Comm. Overhead  \\ \midrule
    Async. FedAvg & 62.47 $\pm$ \textcolor{red}{\textbf{8.20}} & 787.5s & +33.33\% & +25.00\% \\
    Async. Fed-Weighted-Avg & 59.30 $\pm$ \textcolor{red}{\textbf{10.44}} & 1181.3s & +100\% & +87.50\%\\
    Async. FedProx & 59.91 $\pm$ \textcolor{red}{\textbf{7.32}} & 1050.0s & +77.78\% & +66.67\%\\
    Async FjORD & 21.98 $\pm$ \textcolor{red}{\textbf{9.55}} & N/A & N/A & N/A \\
    FedBuff & 57.03 $\pm$ \textcolor{red}{\textbf{11.37}} & 1845.0s & +212.38.85\% & +150.20\%\\
    \midrule 
    \texttt{Hetero AsyncDrop} & \textcolor{teal}{\textbf{66.89}} $\pm$ \textcolor{red}{\textbf{5.36}}  & \textcolor{teal}{\textbf{590.6s}} & \textcolor{teal}{\textbf{[Best]}} & \textcolor{teal}{\textbf{[Best]}} \\ 
    \texttt{AsyncDrop} & 60.02 $\pm$ \textcolor{red}{\textbf{10.38}}  & 787.5s & +33.33\% & +33.33\% \\
    \bottomrule
\end{tabular}
\end{footnotesize}
\label{cifar10_non_iid_final_results}
\end{table*}

\textcolor{teal}{\textbf{\texttt{Hetero \!\!\!\!\! AsyncDrop}: its ingredients}}. 
The above are encapsulated with a weight score function $v(\cdot)$, a device-capacity score function $\psi(\cdot)$ and the mask generator function $\varphi(\cdot)$ in Algorithm \ref{alg:async_IST_smart}.
The function $v(\cdot)$ quantifies the update speed of each weight. 
One simple score can be the norm of weight change: Given predefined grouping of weights in set $\mathcal{J}$, define $\mathbf{W}_{t}^j$ for $j \in \mathcal{J}$ as the $j$-th weight/group of weights of global network at the $t$-th update.
Then, the score function is selected as: 
$v(\mathbf{W}_{t}^j)=\norm{\mathbf{W}_{t}^j-\mathbf{W}_{0}^j}_1$,
i.e., the score function measures \emph{how far from the initial values the $j$-th group of weights has moved.}
In our experiments, we have tested grouping of weights $\mathcal{J}$ by filters and by layers.
Further, we have some measure of the computation capacity of each device $\psi(i)$, and we order the devices in a descending order with respect to capabilities.
The mask generator function $\varphi(\mathbf{W}, \psi(i), v(\cdot))$ generates masks that nullify weights in $\mathbf{W}$, based on the score function $v$ over the weights, and the device capacity list in $\psi(\cdot)$. 
The \texttt{Hetero \!\!\!\!\! AsyncDrop} strategy is that for the $i$-th fastest worker, we drop weights with $i$-th largest $q$ score. 

\vspace{-0.2cm}
\section{Experiments}
\label{Sec:Experiments}
\vspace{-0.2cm}

\textcolor{teal}{\textbf{Setup.}}
We generate simulated FL scenarios with 104 clients/devices of diverse computation and communication capabilities. 
We implement clients as independent processes, each distributed on different GPUs with access to the same RAM space. 
We follow HogWild!'s distributed model \cite{https://doi.org/10.48550/arxiv.1106.5730}: $i)$ we use a shared-memory system to store the global model; $ii)$ each simulated client can update/read the global model in a fully lock free mode; and $iii)$ each client transfers the local model to the assigned GPU for local training. 
We activate 8 clients at any given moment.

We use 25\% dropout rate in (\texttt{Hetero}) \texttt{AsyncDrop} for CNN and MLP, while we use 12.5\% for LSTM. \textit{Even for such low dropout rates, the gains in training are obvious and significant, as we show in the experiments.}
In the appendix, we provide ablation studies on how the dropout rate affects the performance of \texttt{AsyncDrop}-family of algorithms.
We simulate the communication and computation savings by inserting shorter time delay, based on which we estimate the training time and communication cost. 

\textcolor{teal}{\textbf{Simulation of heterogeneous computations.}}
We abstract heterogeneous computation and communication capabilities by ``forcing'' different delays after each training iteration. 
The delay time is inverse proportional to the intended capacity. 
In our experiments, we simulated 8 levels of computation and communication capabilities, that are evenly distributed between the slowest client and fastest clients.
The difference between the slowest and the fastest clients is selected to be $\sim 5\times$. 
We make sure that, at any given moment, clients with diverse capacity are active. 
Finally, all clients with similar computation power shall have similarly biased local data distribution.\footnote{This setting is to avoid the fastest clients with similar computation power cover all the data, which will reduce the problem into a trivial synchronous federated learning problem.}

\textcolor{teal}{\textbf{Problem cases.}}
We experiment on diverse neural network architectures and diverse types of learning tasks, including ResNets on Computer Vision datasets (CIFAR10, CIFAR100, FMNIST), MLPs on FMNIST dataset, and LSTMs on sentimental analysis (IMDB).

\begin{figure*}[!ht]
\vspace{0.1cm}
\centering
\includegraphics[width=0.95\linewidth]{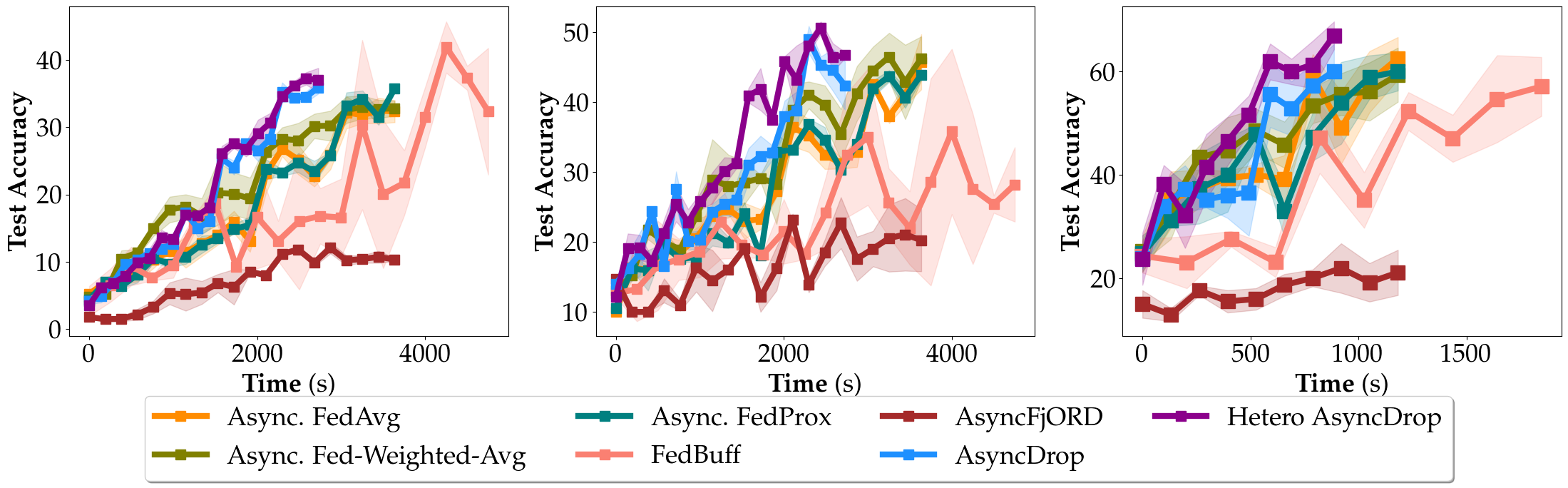}
\vspace{-0.5cm}
\caption{ResNet-based model. \textit{\textbf{Left:}} CIFAR100 non-i.i.d.; \textit{\textbf{Middle:}} CIFAR10, non-i.i.d.; \textit{\textbf{Right:}} FMNIST, non-i.i.d.}
\label{fig:async_all_combine}
\end{figure*}

\textcolor{teal}{\textbf{Baseline methods.}}
For comparison, the baselines we consider are: $i)$ the asynchronous FedAvg is the direct adaptation of FedAvg with asynchronous motions; $ii)$ the asynchronous FedAvg with weighted aggregation represents the general approach of assigning devices different ``importance'', based on their capabilities \cite{https://doi.org/10.48550/arxiv.1903.03934, chen2019communication, 9407951}; $iii)$ the asynchronous FjORD is our asynchronous adaptation of \cite{fjord}; $iv)$ the asynchronous FedProx adds an independent proximal loss to the local training loss of each device, in order to control gradient staleness and overfiting \cite{https://doi.org/10.48550/arxiv.1812.06127}; and $v)$ FedBuff is a semi-asynchronous method which uses buffers for stale updates in a synchronized global scheme \cite{nguyen2022federated}.

For all baselines and (\texttt{Hetero}) \texttt{AsyncDrop} on CIFAR10, CIFAR100, FMNIST, we set the local iterations at $l=50$ while for IMDB, $l=40$. For FedBuff, we set the buffer size to 4, which is half of the activated clients.
We perform 3 trials with different random seeds. 
We report the maximum test accuracy, as well as the estimated time and communication cost to reach a certain target accuracy: 
\textit{we select the second lowest test accuracy among all baselines as the target accuracy (third column in result tables).}


\textcolor{teal}{\textbf{CNN-based results.}}
We test (\texttt{Hetero}) \texttt{AsyncDrop} on ResNet34 and using CIFAR10, CIFAR100 and FMNIST datasets. 
For the CIFAR10 and CIFAR100 datasets, we train for 320 epochs, while for the FMNIST dataset, we train for 160 epochs. 
We stop the execution when the fastest client finishes all its epochs. 
As shown in Table \ref{cifar10_non_iid_final_results}, \texttt{Hetero} \texttt{AsyncDrop} shows non-trivial improvements in \textit{final accuracy, training time and communication cost, simultaneously}. 
\texttt{Hetero} \texttt{AsyncDrop} shows lower accuracy compared with FedBuff in the CIFAR100 case; yet, it achieves up to $85\%$ reduction in training time, due to the fact that FedBuff will require faster workers to wait until the buffer is filled to update the global model (this also justifies the up to $\sim 42.22\%$ reduction in total communication cost). 
Finally, we observe that (\texttt{Hetero}) \texttt{AsyncDrop} shows quite stable performance; in red color we indicate the variability of results over trials. The similar training time of some baselines to reach target accuracy is caused by epoch-wise testing, using same epoch-wise learning rate schedule and similar convergence rate as shown in Figure \ref{fig:async_all_combine}.

\textcolor{teal}{\textbf{MLP-based and LSTM-based results.}}
We adapt the (\texttt{Hetero}) \texttt{AsyncDrop} to the MLP model by applying the hidden neuron Dropout, which is similar to channel dropout  in CNNs (160 epochs). 
For the LSTM and the IMDB sentimental analysis dataset (80 epochs), we create non-i.i.d. datasets based on different label distribution in each local training set.
As shown in Table 1 in the Appendix, \texttt{Hetero} \texttt{AsyncDrop} achieves better performance overall over the MLP model, in terms of accuracy, training time and communication cost. 
As shown in Table 2 in the Appendix, for the LSTM model, \texttt{Hetero AsyncDrop} shows comparable accuracy with respect to other baselines, while achieving reduction in training time and communication cost in most cases. Our conjecture for the lower gain compared with other architectures is that LSTM-based (or even RNN-based) architectures might be difficult to our proposed dropout score mechanism, as the update of each network parameter is the average of several virtual parameters in the unrolled network.

\vspace{-0.2cm}
\section{Concluding Remarks}
\label{Sec:Conclusion}
\vspace{-0.2cm}
We present (\texttt{Hetero}) \texttt{AsyncDrop}, a novel algorithm for asynchronous distributed neural network training in the FL scenario. \texttt{AsyncDrop} operates by asynchronously creating submodels out of the global model, in order to train those independently on heterogeneous devices. These models are asynchronously aggregated into the global model. By only communicating and training submodels over edge workers, \texttt{AsyncDrop} reduces the communication and local training cost. We demonstrate the impact of \texttt{AsyncDrop} on MLPs, ResNets and LSTMs over non-i.i.d. distributions of CIFAR10, CIFAR100, FMNIST and IMDB datasets. Additional experiments in the appendix include ablation analysis on how dropout ratio, local iteration length and dropout strategy affect the final performance.  

We aim to extend \texttt{AsyncDrop} to other network architectures, such as Transformers. Further, \texttt{AsyncDrop} is fully-compatible with various gradient compression methods, which could potentially further improve the performance. We will investigate the prospect of fully integrating such compression methods within \texttt{AsyncDrop}, both during training and communication phases in future work. 

\paragraph{Acknowledgements.}
This work is supported by NSF FET:Small no. 1907936, NSF MLWiNS CNS no. 2003137 (in collaboration with Intel), NSF
CMMI no. 2037545, NSF CAREER award no. 2145629, a Welch Foundation Research Grant, no. 964181, and Rice InterDisciplinary Excellence Award (IDEA).

\bibliography{biblio.bib}
\bibliographystyle{plain}

\clearpage

\appendix
\onecolumn
\section{Experimental Results using \texttt{AsyncDrop} for the MLP and LSTM architectures}

We adapt the (\texttt{Hetero}) \texttt{AsyncDrop} to the MLP and LSTM model by applying the hidden neuron Dropout, which is similar to channel dropout  in CNNs. 
For the LSTM and the IMDB sentimental analysis dataset, we create non-i.i.d. datasets based on different label distribution in each local training set.
For the FMNIST experiments, we train for 160 epochs, while for the IMDB experiments, we train for 80 epochs.
As shown in Table 1, \texttt{Hetero} \texttt{AsyncDrop} achieves better performance overall over the MLP model, in terms of accuracy, training time and communication cost. 
As shown in Table 2, for the LSTM model, \texttt{Hetero AsyncDrop} shows comparable accuracy with respect to other baselines, while achieving reduction in training time and communication cost in most cases. Our conjecture for the lower gain compared with other architectures is that LSTM-based (or even RNN-based) architectures might be difficult to our proposed dropout score mechanism, as the update of each network parameter is the average of several virtual parameters in the unrolled network.

\begin{figure*}[!ht]
\vspace{0.1cm}
\centering
\includegraphics[width=0.95\linewidth]{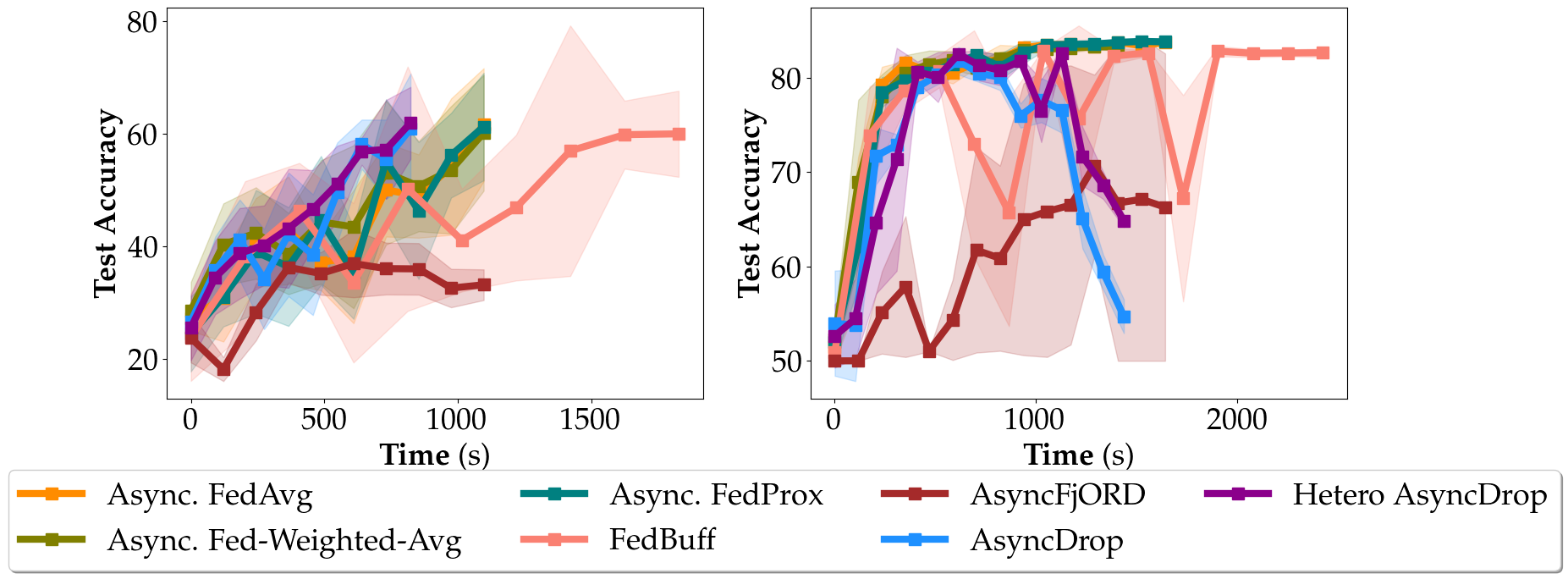}
\vspace{-0.5cm}
\caption{ \textit{\textbf{Left:}} MLP FMNIST non-i.i.d.;  \textit{\textbf{Right:}} LSTM IMDB, non-i.i.d.}
\label{fig:async_all_combine}
\end{figure*}

\begin{table*}[!htp]
\centering
\caption{Test accuracy of asynchronous FL baselines vs. (\texttt{Hetero}) \texttt{AsyncDrop} using a MLP architecture on non-i.i.d MNIST data over $>100$ clients. We report the time and communication overhead to reach a certain target accuracy: 
``Time for \texttt{XX}\% Accuracy'' denotes the second lowest test accuracy among all baselines as the target accuracy.} 
\begin{small}
\begin{tabular}{ccccc}
\toprule
    &   Max. Test Accuracy & Time for 59.95\% Accuracy & Time Overhead & Comm. Overhead  \\ \midrule
    Async. FedAvg & 61.59 $\pm$ \textcolor{red}{\textbf{10.02}} & 1096.1s & +33.33\% & +28.73\% \\
    Async. Fed-Weighted-Avg & 60.14 $\pm$ \textcolor{red}{\textbf{10.28}} & 1096.1s & +33.33\% & +28.73\%\\
    Async. FedProx & 61.25 $\pm$ \textcolor{red}{\textbf{9.54}} & 1096.1s &+33.33\% & +28.73\%\\
    Async. FjORD & 36.92 $\pm$ 6.00 & N/A & N/A & N/A \\
    FedBuff & 59.96 $\pm$ \textcolor{red}{\textbf{7.64}} & 1827.3s & +166.50\% & +83.90\%\\ \midrule
    \texttt{Hetero AsyncDrop} & 61.90 $\pm$ 6.39  & \textcolor{teal}{\textbf{822.5s}} & \textcolor{teal}{\textbf{[Best]}} & \textcolor{teal}{\textbf{[Best]}} \\ 
    \texttt{AsyncDrop} & 60.85 $\pm$ \textcolor{red}{\textbf{9.78}}  & \textcolor{teal}{\textbf{822.5s}} & \textcolor{teal}{\textbf{[Best]}} & \textcolor{teal}{\textbf{[Best]}} \\
    \bottomrule
\end{tabular}
\end{small}
\label{fmnist_mlp_non_iid_final_results}
\end{table*}

\begin{table*}[!htp]
\centering
\caption{Test accuracy of asynchronous FL baselines vs. (\texttt{Hetero}) \texttt{AsyncDrop} using a LSTM architecture on non-i.i.d IMDB data over $>100$ clients. We report the time and communication overhead to reach a certain target accuracy: 
``Time for \texttt{XX}\% Accuracy'' denotes the second lowest test accuracy among all baselines as the target accuracy.} 
\begin{small}
\begin{tabular}{ccccc}
\toprule
    &   Max. Test Accuracy & Time for 82\% Accuracy & Time Overhead & Comm. Overhead  \\ \midrule
    Async. FedAvg & 83.73 $\pm$ 0.18 & 1875.6s & +52.38\% & +32.06\% \\
    Async. Fed-Weighted-Avg & 83.26 $\pm$ 0.33 & 1640.6s & +33.33\% & +15.56\%\\
    Async. FedProx & 83.83 $\pm$ 0.20 & 1406.3 & +14.28\% & \textcolor{teal}{\textbf{-0.95\%}}\\
    Async. FjORD & 70.68 $\pm$ 9.62 & N/A & N/A & N/A \\
    FedBuff & 82.81 $\pm$ 0.30 & 1038.1s & +68.71\% & +14.28\%\\
    \midrule
    \texttt{Hetero AsyncDrop} & 82.60 $\pm$ 0.57  & \textcolor{teal}{\textbf{1230.4s}} & \textcolor{teal}{\textbf{[Best]}} & -\\ 
    \texttt{AsyncDrop} & 81.70 $\pm$ 1.45  & N/A & N/A & N/A \\
 \bottomrule
\end{tabular}
\end{small}
\label{imdb_lstm_non_iid_final_results}
\end{table*}

\section{Ablation Study}
\textcolor{teal}{\textbf{Dropout Ratio for \texttt{Hetero AsyncDrop}.}}
In this section we analyze how dropout ratio affects the final accuracy and training time. 
As shown in Table \ref{ablation_1}, low dropout rates result in longer training time and lower final accuracy. 
Lower final accuracy is mainly due to the fact that \texttt{AsyncDrop} with dropout work as a regularization technique that works favorably towards better final accuracy.
On the other hand, higher dropout rates  cause relative reduction in training time, but significant reduction in the final accuracy. This shows dropping too many parameters causes the training of local model unstable, reducing the final accuracy of the global model.

\begin{table}[!htp]
\centering
\caption{Test accuracy of \texttt{Hetero AsyncDrop} with different dropout rate on non-i.i.d. CIFAR100 datasets.} 

\begin{tabular}{cccc}
\toprule
     Dropout Ratio &   0\% & 25\% & 50\%  \\ \midrule
    Test Accuracy & 32.47 $\pm$ 1.89  & 37.26 $\pm$ 0.93 & 23.95 $\pm$ 1.07 \\
    Time for Accuracy to 23\% & 2105.4s  & 1579.1s & 1531.2s \\
    
 \bottomrule
\end{tabular}
\label{ablation_1}
\end{table}

\begin{figure*}[!ht]
\vspace{0.1cm}
\centering
\includegraphics[width=0.95\linewidth]{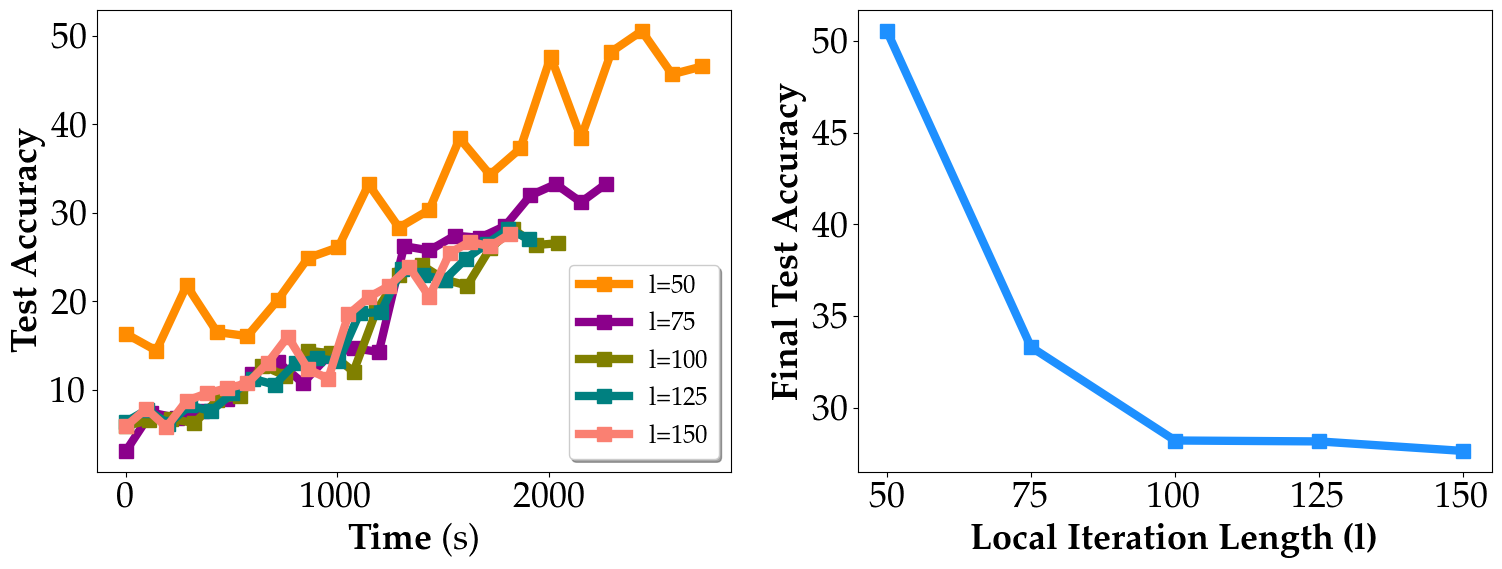}
\vspace{-0.5cm}
\caption{Test Accuracy of \texttt{Hetero AsyncDrop} with different local iteration length on CIFAR100, non i.i.d}
\label{fig:ablation_local_iteration}
\end{figure*}

\textcolor{teal}{\textbf{Local Iteration Length for \texttt{Hetero AsyncDrop}}.} In this section we analyze how the number of local iterations ($l$) affect the final accuracy and training time. Ideally, longer local training iteration will decrease the communication frequency and, thus, the communication and time cost. \cite{stich2018local,dun2022resist} However, as shown in Figure \ref{fig:ablation_local_iteration}, increasing local training iterations cause significant decrease in final accuracy, which might be due to $i)$ increased gradient staleness from slower clients; $ii)$ more severe model drifting. In future research, we will study how to improve \texttt{Hetero AsyncDrop} in order to be more robust for higher $l$ values.

\textcolor{teal}{\textbf{The effectiveness of Smart Partition in \texttt{Hetero AsyncDrop}}.} As shown in Table \ref{ablation_2}, on all datasets and all network architectures, \texttt{Hetero AsyncDrop} provides improvements on the final accuracy, while decreasing the variance.
This indicates that \texttt{Hetero AsyncDrop} could be a more stable method that balances heterogeneous clients and data. 
By interpreting current experiments, \texttt{Hetero AsyncDrop} gives the most gain on ResNet (or CNN model family), which utilizes a channel dropout method. 

On the other hand, in both MLP and LSTM, we dropout output hidden neurons and the network parameters accordingly. As filter channels in ResNet are considered to be more semantically disentangled from each other, compared to hidden neurons in fully connected layers, \texttt{Hetero AsyncDrop}'s score for filters in ResNet capture better which filters are more frequently/significantly updated (and thus biased towards the faster clients). 
Thus, our scoring function $q(\cdot)$ targets which parameters to drop in ResNet architectures. We leave to future research on how to improve our proposed mechanism on MLP and LSTM architectures, potentially by using a more structured dropout strategy.

\begin{table}[!h]
\centering
\caption{Test accuracy of \texttt{Hetero AsyncDrop} versus \texttt{AsyncDrop}} 
\begin{tabular}{cccc}
\toprule
      &  \texttt{Hetero AsyncDrop} & \texttt{AsyncDrop} & Relative Accuracy Gain \\ \midrule
    ResNet+CIFAR10 & 50.67 $\pm$ 1.75  & 48.98 $\pm$ 3.87 & +3.45\% \\
    ResNet+CIFAR100 & 37.26 $\pm$ 0.93 & 35.93 $\pm$ 1.07 & +2.98\% \\
    ResNet+FMNIST & 66.89 $\pm$ 5.36 & 60.02 $\pm$ 10.38 & +11.45\% \\
    MLP+FMNIST & 61.90 $\pm$ 6.39 & 60.85 $\pm$ 9.78 & +1.72\% \\
    LSTM+IMDB & 82.60 $\pm$ 0.57 & 81.70 $\pm$ 1.45 & +1.10\% \\
 \bottomrule
\end{tabular}
\label{ablation_2}
\end{table}

\begin{figure*}[!ht]
\vspace{0.1cm}
\centering
\includegraphics[width=0.65\linewidth]{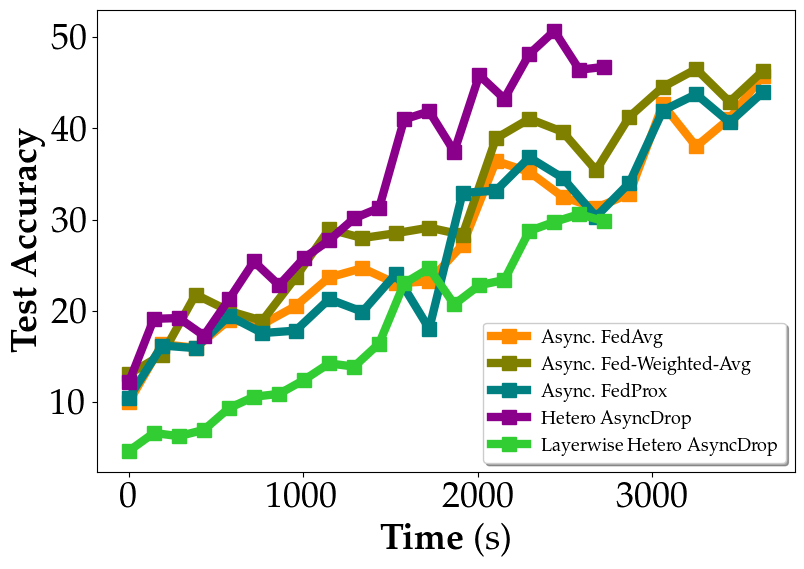}
\vspace{-0.5cm}
\caption{Test Accuracy of Layerwise \texttt{Hetero AsyncDrop} versus all baselines on CIFAR100 non- i.i.d.}
\label{fig:ablation_layerwise}
\end{figure*}

\textcolor{teal}{\textbf{Layerwise Smart Partition for \texttt{AsyncDrop} on ResNet}.} 
For ResNet architectures, another commonly used dropout strategy is that of layerwise dropout \cite{huang2016deep,dun2022resist}.
We adapt \texttt{Hetero AsynDrop} using layerwise Smart Parition and dropout. 
The main difference from original \texttt{Hetero AsyncDrop} algorithm is that we consider each layer as a single unit for dropout. 
The Smart Partition score function calculates the total norm change of all parameters in each layer and ranks layers accordingly. 
Finally, for the $i$-th fastest worker, we drop layers with the $i$-th largest $q$ score. In experiments, we set the dropout rate to be 25\% to keep the submodel size similar to the original \texttt{AsyncDrop} experimebts. 
As shown in Figure \ref{fig:ablation_layerwise}, Layerwise \texttt{Hetero AsyncDrop} results in a non-negligible decease in performance, with respect to the final accuracy. This might show layer dropping causes increased variance and instability, especially in asynchronous training.

\section{Detailed Mathematical Formulation of \texttt{AsyncDrop}}
\label{detailed_math_form}
For a vector $\mathbf{v}$, $\norm{\mathbf{v}}_2$ denotes its Euclidean ($\ell_2$) norm. For a matrix $\mathbf{V}$, $\norm{\mathbf{V}}_F$ denotes its Frobenius norm. We use $\mathcal{P}\paren{\cdot}$ to denote the probability of an event, and $\indy{\cdot}$ to denote the indicator function. For two vectors $\mathbf{v}_1,\mathbf{v}_2$, we use the simplified notation $\indy{\mathbf{v}_1;\mathbf{v}_2} := \indy{\inner{\mathbf{v}_1}{\mathbf{v}_2}\geq 0}$. Given that the mask in iteration $t$ is $\mathbf{M}_t$, we denote $\E_{[\mathbf{M}_t]}\left[\cdot\right] = \E_{\mathbf{M}_0,\dots,\mathbf{M}_t}\left[\cdot\right]$.

Recall that the CNN considered in this paper has the form:
\begin{align*}
    f(\mathbf{x},\theta) = \inner{\mathbf{a}}{\sigma\paren{\W_1\hat{\phi}\paren{\x}}}.
\end{align*}
Denote $\hat{\x} = \hat{\phi}\paren{\x}$. Essentially, this patching operator applies to each channel, with the effect of extending each pixel to a set of pixels around it. So we denote $\hat{\x}_i^{(j)}\in \R^{q\hat{d}}$ as the extended $j$th pixel across all channels in the $i$th sample. For each transformed sample, we have that $\norm{\hat{\x}_i}_F\leq \sqrt{q}\norm{\x_i}_F$. We simplify the CNN output as:
\begin{align*}
    f(\hat{\x},\W) = \inner{\mathbf{a}}{\sigma\paren{\W\otimes\hat{\x}}} = \sum_{r=1}^{m_1}\sum_{j=1}^pa_{rj}\sigma\paren{\inner{\hat{\x}^{(j)}}{\w_r}}.
\end{align*}
In this way, the formulation of CNN reduces to MLP despite a different form of input data $\hat{\x}$ and an additional dimension of aggregation in the second layer.
We consider training the neural network $f$ on the mean squared error (MSE):
\begin{align*}
    \mathcal{L}\paren{\W} = \norm{f\paren{\hat{\X},\W} - \y}_2^2 = \sum_{i=1}^n\paren{f\paren{\hat{\x}_i,\W} - y_i}^2.
\end{align*}

Similar to HogWild! training is fully lock free and thus cannot prevent several clients from updating the global model at same time, in order to be closer to real-world case, in theoretical analysis, we extend \texttt{AsyncDrop} by allowing $S\geq 1$ clients --with same computation speed and thus same global model update schedule-- to update the global model simultaneously by averaging their updates. 
This can be reduced back to original \texttt{AsyncDrop}, by simply setting $S=1$ without changing any part of the analysis. 

We assume that we can group all clients by their computation/communication power, such that, in each group, there are exactly $S\geq1$ clients with the same power; thus, with the same global model update schedule. 
Each worker will be independently assigned to different dropout masks to be applied on the global model.
All updates from local clients in the same group will be averaged. 

In math, the subnetwork on client $s$ by filter-wise partition is given by:
\begin{align*}
    f_{\mathbf{m}^{(s)}}\paren{\hat{\x},\W} = \sum_{r=1}^{m_1}\sum_{j=1}^pm^{(s)}_ra_{rj}\sigma\paren{\inner{\hat{\x}^{(j)}}{\w_r}}.
\end{align*}
Trained on the regression loss, the surrogate gradient on current network parameters is given by:
\begin{align*}
    \nabla_{\w_r}\mathcal{L}_{\mathbf{m}^{(s)}}\paren{\W} = m^{(s)}_r \sum_{i=1}^n\sum_{j=1}^p\paren{f_{\mathbf{m}^{(s)}}\paren{\hat{\x}_i,\W} - y_i}a_{rj}\hat{\x}_{i}^{(j)}\indy{\hat{\x}_i^{(j)};\w_r}.
\end{align*}
We correspondingly scale the whole network function:
\begin{align*}
    f\paren{\hat{\x},\W} = \xi\sum_{r=1}^m\sum_{j=1}^pa_{rj}\sigma\paren{\inner{\hat{\x}_i^{(j)}}{\w_r}}.
\end{align*}
Assuming it is training on the MSE, we write out its gradient as:
\begin{align*}
    \nabla_{\w_r}\mathcal{L}\paren{\W} = \xi\sum_{i=1}^n\sum_{j=1}^n\paren{f\paren{\hat{\x}_i,\W} - y_i}a_{rj}\hat{\x}_i^{(j)}\indy{\hat{\x}_i^{(j)};\w_r}.
\end{align*}
In this work, we consider the \textsc{AsynDrop} with one step of local iteration training and $\delta_t\leq E$ gradient delay caused by asynchronous training, given by:
\begin{align*}
    \w_{r,t+1} = \w_{r,t} - \eta\frac{N_{r,t-\delta_t}^\perp}{N_{r,t-\delta_t}}\sum_{s=1}^S\nabla_{\w_r}\mathcal{L}_{\mathbf{m}_{t-\delta_t}^{(s)}}\paren{\W_{r,t-\delta_t}}.
\end{align*}
Here, let $N_{r,t} = \max\left\{\sum_{s=1}^Sm_{r,t}^{(s)},1\right\}$, and $N_{r,t}^\perp = \min\left\{\sum_{s=1}^Sm_{r,t}^{(s)},1\right\}$. Intuitively, $N_{r,t}$ denote the "normalizer" that we will divide the sum of the gradients from all subnetworks with, and $N_{r,t}^\perp$ denote the indicator of whether filter $r$ is trained in at least one subnetwork. Let $\theta = \mathcal{P}\paren{N_{r,t}^\perp = 1} = 1 - (1 - \xi)^p$, denoting the probability that at least one of $\left\{m_{r,t}^{(s)}\right\}_{s=1}^S$ is one. Denote $u_t^{(i)} = f\paren{\hat{\x}_i,\W_t}$. For further convenience of our analysis, we define
\begin{align*}
    \tilde{u}_{r,t}^{(i)} = \frac{N_{r,t}^\perp}{N_{r,t}}\sum_{s=1}^Sm_{r,t}^{(s)}\hat{u}_t^{(s, i)};\quad \mathbf{g}_{r,t} = \frac{N_{r,t}^\perp}{N_{r,t}}\sum_{s=1}^S\nabla_{\w_r}\mathcal{L}_{\mathbf{m}_t^{(s)}}\paren{\W_t}.
\end{align*}
Then the \texttt{AsyncDrop} training has the form
\begin{align*}
    \w_{r,t+1} = \w_{r,t} - \eta\mathbf{g}_{r,t-\delta_t};\quad
    \mathbf{g}_{r,t} = \sum_{i=1}^n\sum_{j=1}^pa_{rj}\paren{\tilde{u}_{r,t}^{(i)} - N_{r,t}^\perp y_i}\hat{\x}_i^{(j)}\indy{\hat{\x}_i^{(j)};\w_{r,t}}.
\end{align*}
Suppose that Assumptions 1 and 2 in main text  hold. Then for all $i\in[n]$ we have $\norm{\x_i}_F = q^{\frac{1}{2}}$, and for all $i,i'\in[n]$ such that $i\neq i'$, we have $\x_i\neq \x_{i'}$.  As in previous work \cite{du2019gradient}, we have $\norm{\hat{\x}_i}_F  = \leq \sqrt{q}\norm{\x_i}_F \leq 1$. Thus, for all $j\in[p]$ we have $\norm{\hat{\x}_i^{(j)}}\leq 1$. Moreover, since $\x_i\neq \x_{i'}$ for $i\neq i'$, we then have $\hat{\x}_i\neq \hat{\x}_{i'}$, which implies that $\hat{\x}_i^{(j)}\neq \hat{\x}_{i'}^{(j)}$ for all $i\neq i'$ and $j\in[p]$.
\section{Proof of \texttt{AsyncDrop} Convergence}

\label{convergence_proof}
In this section, our goal is to prove the convergence of extended \texttt{AsyncDrop}. We first state the extended version here, and proceed proving it.

\begin{thm}
\label{theorm_1}
\textit{Let $f(\cdot, \cdot)$ be a one-hidden-layer CNN with the second layer weight fixed. 
Let $\mathbf{u}_t$ abstractly represent the output of the model after $t$ iterations, over the random selection of the masks.
Let $E$ denotes the maximum gradient delay/staleness. 
Let $\xi$ denote the dropout rate ($\xi = 1$ dictates that all neurons are selected), and denote $\theta = 1 - (1 - \xi)^S$ the probability that a neuron is active in at least one subnetwork.
Assume the number of hidden neurons satisfies
$m = \Omega\paren{\max\{\frac{n^4K^2}{\lambda_0^4\delta^2}\max\{n, d\},\frac{n}{\lambda_0}\}}$ and the step size satisfies $\eta = O\paren{\frac{\lambda_0}{n^2}}$.
Let $\kappa$ be a proper initialization scaling factor, and it is considered constant.
We use $\lambda_0$ to denote the smallest eigenvalue of the Neural Tangent Kernel matrix.
Let Assumptions 1 and 2 be satisfied. Then, the following convergence rate guarantee is proved to be satisfied:}
\end{thm}
\begin{align*}
\E_{\mathbf{M}_t}\left[\norm{\U_{t+1} - \y}_2^2\right] & \leq \textcolor{red}{\paren{1 - \frac{\theta\eta\lambda_0}{4}}^{t}\norm{\U_0- \y}_2^2} \\
    & \quad\quad\quad + O\left(\textcolor{violet}{\frac{\theta\eta\lambda_0^3 \xi^2 \kappa^2 E^2}{n^2}} + \textcolor{orange}{\frac{\xi^2(1-\xi)^2\theta\eta n^3\kappa^2d}{m\lambda_0}} + \textcolor{violet}{\frac{\eta^2\theta^2 n \kappa ^2 \lambda_0 \xi^4 E^2}{m^{4}}} + \textcolor{orange}{\frac{\xi^2(1-\xi)^2\theta^2\eta^2 n^2  \kappa^2 d}{m^3\lambda_0 }} \right. \\
    &\left. \quad\quad\quad\quad\quad + \textcolor{violet}{\frac{\xi^2(1-\xi)^2\theta^2\eta^2 \kappa^2 \lambda_0 E^2}{m^3}} + \textcolor{orange}{\frac{\xi^2(1-\xi)^2\theta^2\eta^2 n^2  \kappa^2 d}{m^2\lambda_0}}+\textcolor{orange}{\frac{n\kappa^2\paren{\theta-\xi^2}}{S}}\right).
\end{align*}
\vspace{-0.5cm}

We care about the MSE computed on the scaled full network:
\begin{align*}
    u_k^{(i)} = \xi\sum_{r=1}^m\sum_{j=1}^pa_{rj}\sigma\paren{\inner{\hat{\x}_i^{(j)}}{\w_{r,t}}};\quad\mathcal{L}\paren{\W_t} = \norm{\U_t - \y}_2^2.
\end{align*}
Performing gradient descent on this scaled full network involves computing:
\begin{align*}
    \nabla_{\w_r}\mathcal{L}\paren{\W_t} = \xi\sum_{i=1}^n\sum_{j=1}^p\paren{u_t^{(i)} - y_i}a_{rj}\xij\indy{\xij;\w_{r,t}}.
\end{align*}
We will prove Theorem \ref{theorm_1} by induction. We assume therom \ref{theorm_1} is true for all $t' < t$ and $\norm{\w_{r,t'} - \w_{r,0}}_2\leq R := O\paren{\frac{\kappa\lambda_0}{n}}$.

\subsection{Change of Activation Pattern}
Let $R$ be some fixed scale. For convenience, we denote:
\begin{align*}
    A_{ir}^{(j)} = \left\{\exists\w: \norm{\w - \w_{r,0}}_2\leq R; \indy{\hat{\x}_i^{(j)};\w}\neq\indy{\hat{\x}_i^{(j)};\w_{r,0}}\right\}.
\end{align*}
Note that $A_{ir}^{(j)}$ happens if and only if $\left|\inner{\hat{\x}_i^{(j)}}{\w_{r,0}}\right| < R$. Therefore $\mathbb{P}\paren{A_{ir}^{(j)}} < \frac{2R}{\kappa\sqrt{2\pi}}$. Denote:
\begin{align*}
    P_{ij} = \left\{r\in[m]:\neg A_{ir}^{(j)}\right\};\quad P_{ij}^\perp = [m]\setminus P_{ij}.
\end{align*}
The next lemma shows the magnitude of $P_{ij}^\perp$ is upper bounded by a controlled quantity.

\begin{lem}
Let $m = \Omega\paren{R^{-1}\log\frac{np}{\delta}}$. Then with probability at least $1-O(\delta)$ it holds for all $i\in[n]$ and $j\in[p]$ that
\begin{align*}
    \left|P_{ij}^\perp\right| \leq 3m\kappa^{-1}R.
\end{align*}
\end{lem}

\textit{Proof:}
The magnitude of $P_{ij}^\perp$ satisfies:
\begin{align*}
    \left|P_{ij}^\perp\right| = \sum_{r=1}^m\indy{A_{ir}^{(j)}}.
\end{align*}
The indicator function $\indy{A_{ir}^{(j)}}$ has bounded first and second moment:
\begin{align*}
    \E_{\W}\left[\indy{A_{ir}^{(j)}}\right] = \mathbb{P}\paren{A_{ir}^{(j)}} & \leq \frac{2R}{\kappa\sqrt{2\pi}}, \\ \E_{\W}\left[\paren{\indy{A_{ir}^{(j)}-\E_{\W}\left[\indy{A_{ir}^{(j)}}\right]}}^2\right]& \leq \E_{\W}\left[\indy{A_{ir}^{(j)}}^2\right] \leq \frac{2R}{\kappa\sqrt{2\pi}}.
\end{align*}
This allows us to apply the Berstein Inequality to get that:
\begin{align*}
    \mathbb{P}\paren{\sum_{r=1}^m\indy{A_{ir}^{(j)}} > \frac{2mR}{\kappa\sqrt{2\pi}} + mt} < \exp\paren{-\frac{m\kappa t^2\sqrt{2\pi}}{8\paren{1 + \frac{t}{3}}R}}.
\end{align*}
Therefore, with probability at least $1 - np\exp\paren{-m\kappa^{-1} R}$ it holds for all $i\in[n]$ and $j\in[p]$ that:
\begin{align*}
    \left|P_{ij}^\perp\right| = \sum_{r=1}^m\indy{A_{ir}^{(j)}} \leq 3m\kappa^{-1}R.
\end{align*}
Letting $m = \Omega\paren{R^{-1}\log\frac{np}{\delta}}$ gives that the success probability is at least $1 - O(\delta)$.

\subsection{Initialization Scale}Let $\w_{0,r}\sim\mathcal{N}\paren{0, \kappa^2\mathbf{I}}$ and $a_{j}\sim\left\{-\frac{1}{p\sqrt{m}},\frac{1}{p\sqrt{m}}\right\}$ for all $r\in[m]$ and $j\in[p]$. The following lemmas hold true: 
\begin{lem}
\label{init_mat_bound}
Suppose $\kappa \leq 1, R\leq \kappa\sqrt{\frac{d}{32}}$. With probability at least $1 - e^{md/32}$ we have that:
\begin{align*}
    \|W_0\|_F \leq  \kappa\sqrt{2md} - \sqrt{m}R.
\end{align*}
\end{lem}

\begin{lem}
\label{init_inner_prod_bound}
Assume $\kappa \leq 1$ and $R \leq \frac{\kappa}{\sqrt{2}}$. With probability at least $1 - ne^{-\frac{m}{32}}$ over initialization, it holds for all $i\in[n]$ that:
\begin{align*}
    \sum_{r=1}^m\inner{\mathbf{w}_{0,r}}{\mathbf{x}_i}^2 \leq 2m\kappa^2 - mR^2\\
    \sum_{i=1}^n\sum_{r=1}^m\inner{\mathbf{w}_{0,r}}{\mathbf{x}_i}^2 \leq 2mn\kappa^2 - mnR^2.
\end{align*}
\end{lem}
Moreover, we can bound the initial MSE, based on the lemma below:
\begin{lem}
\label{initial_scale}
Assume that for all $i\in[n]$, $y_i$ satisfies $|y_i|\leq C$ for some $C > 0$. Then, we have
\begin{align*}
    \E_{\mathbf{W}_0,\hat{\mathbf{a}}}\left[\|\y - \U_0\|_2^2\right] \leq \paren{p^{-1} + C^2}n.
\end{align*}
\textit{Proof:}
It is obvious that $\E_{\mathbf{W}_0,\hat{\mathbf{a}}}\left[u_0^{(i)}\right] = 0$ for all $i\in[n]$. Moreover:
\begin{align*}
    \E_{\mathbf{W}_0,\hat{\mathbf{a}}}\left[u_0^{(i)2}\right] & = \sum_{r,r'=1}^m\sum_{j,j'=1}^p\E_{\hat{\mathbf{a}}}\left[a_{rj}a_{r'j'}\right]\E_{\W_0}\left[\sigma\paren{\inner{\hat{\x}_{i}^{(j)}}{\w_{0,r}}}\sigma\paren{\inner{\hat{\x}_{i}^{(j')}}{\w_{0,r}}}\right]\\
    & = \frac{1}{p^2m}\sum_{r=1}^m\sum_{j=1}^p\E_{\W_0}\left[\sigma\paren{\inner{\hat{\x}_{i}^{(j)}}{\w_{0,r}}}^2\right]\\
    & \leq \frac{1}{p^2m}\sum_{r=1}^m\sum_{j=1}^p\E_{\W_0}\left[\inner{\hat{\x}_{i}^{(j)}}{\w_{0,r}}\right]\\
    & \leq p^{-1}
\end{align*}
Therefore:
\begin{align*}
    \E_{\mathbf{W}_0,\hat{\mathbf{a}}}\left[\norm{\U_0 -\y}_2^2\right] = \sum_{i=1}^n\E_{\mathbf{W}_0,\hat{\mathbf{a}}}\left[\paren{u_0^{(i)} -y_i}_2^2\right] = \sum_{i=1}^n\paren{\E_{\mathbf{W}_0,\hat{\mathbf{a}}}\left[u_0^{(i)2}\right] + y_i^2} \leq\paren{p^{-1}+ C^2}n
\end{align*}

\end{lem}
\subsection{Kernel Analysis}
The neural tangent kernel is defined to be the inner product of the gradient with respect to the neural network output. We let the finite-width NTK be defined as:
\begin{align*}
    \h(t)_{ii'} = \sum_{r=1}^m\sum_{j,j'=1}^pa_{rj}a_{rj'}\inner{\hat{\x}_i^{(j)}}{\hat{\x}_{i'}^{(j')}}\indy{\hat{\x}_i^{(j)};\w_{r,t}}\indy{\hat{\x}_{i'}^{(j')};\w_{r,t}}.
\end{align*}
Moreover, let the infinite width NTK be defined as
\begin{align*}
    \h^\infty_{ii'} =  \frac{1}{p^2}\sum_{j=1}^p\inner{\hat{\x}_i^{(j)}}{\hat{\x}_{i'}^{(j)}}\E_{\mathbf{w}\sim\mathcal{N}\paren{0,\mathbf{I}}}\left[\indy{\hat{\x}_i^{(j)};\w}\indy{\hat{\x}_{i'}^{(j)};\w}\right].
\end{align*}
Let $\lambda_0 = \lambda_{\min}\paren{\h^\infty}$. Note that since $\hat{\x}_i\not\parallel\hat{\x}_{i'}$ for $i\neq i'$. Thus $\hat{\x}_i^{(j)}\not\parallel\hat{\x}_{i'}^{(j)}$ for $i\neq i'$. () shows that the matrix $\hat{\h}(j)$ , as defined below, is positive definite for all $j\in[p]$:
\begin{align*}
    \hat{\h}(j)_{ii'}^\infty = \inner{\hat{\x}_i^{(j)}}{\hat{\x}_{i'}^{(j)}}\E_{\mathbf{w}\sim\mathcal{N}\paren{0,\mathbf{I}}}\left[\indy{\hat{\x}_i^{(j)};\w}\indy{\hat{\x}_{i'}^{(j)};\w}\right].
\end{align*}
Since $\h^\infty = p^{-2}\sum_{j=1}^p\hat{\h}(j)^\infty$, we have that $\h^\infty$ is positive definite and thus $\lambda_0 > 0$. The following lemma shows that the NTK remains positive definite throughout training.
\begin{lem}
Let $m = \Omega\paren{\lambda_0^{-2}n^2\log\frac{n}{\delta}}$. If for all $r\in[m]$ and all $t$ we have $\norm{\w_{r,t} - \w_{r,0}}_2\leq R := O\paren{\frac{\kappa\lambda_0}{n}}$. Then with probability at least $1-\delta$ we have that for all $t$:
\begin{align*}
    \lambda_{\min}\paren{\h(t)}\geq \frac{\lambda_0}{2}.
\end{align*}
\end{lem}
\textit{Proof:}
To start, we notice that for all $r\in[m]$:
\begin{align*}
    \E_{\mathbf{W}_0,\mathbf{a}}&\left[\sum_{j,j'=1}^pa_{rj}a_{rj'}\indy{\hat{\x}_i^{(j)};\w_{r,0}}\indy{\hat{\x}_{i'}^{(j')};\w_{r,0}}\right] \\
    &\quad\quad\quad= \frac{1}{p^2m}\E_{\w\sim\mathcal{N}(0,\mathbf{I})}\left[\indy{\hat{\x}_i^{(j)};\w}\indy{\hat{\x}_{i'}^{(j)};\w}\right]
\end{align*}
Moreover, we have that:
\begin{align*}
    \left|\sum_{j,j'=1}^pa_{rj}a_{rj'}\inner{\hat{\x}_i^{(j)}}{\hat{\x}_{i'}^{(j')}}\indy{\hat{\x}_i^{(j)};\w_{r,0}}\indy{\hat{\x}_{i'}^{(j')};\w_{r,0}} \right|\leq 1.
\end{align*}
Thus, we can apply Hoeffding's inequality with bounded random variable to get that
\begin{align*}
    \mathbb{P}\paren{\left|\h(0)_{i,i'} - \h^\infty_{i,i'}\right| \geq t}\leq 2\exp\paren{-mt^2}.
\end{align*}
Therefore, with probability at least $1-O(\delta)$ it holds that for all $i,i'\in[n]$:
\begin{align*}
    \left|\h(0)_{i,i'} - \h^\infty_{i,i'}\right|\leq \frac{\log\frac{n}{\delta}}{\sqrt{m}},
\end{align*}
which implies that:
\begin{align*}
    \norm{\h(0) - \h^\infty}\leq \norm{\h(0) - \h^\infty}_F \leq \frac{n\paren{\log\delta^{-1}+\log n}}{\sqrt{m}}.
\end{align*}
As long as $m = \Omega\paren{\lambda_0^{-2}n^2\log\frac{n}{\delta}}$, we will have:
\begin{align*}
    \norm{\h(t) - \h^\infty}\leq \frac{\lambda_0}{4}.
\end{align*}
Now, we move on to bound $\norm{\h(t) - \h(0)}$. We have that:
\begin{align*}
    \h(t)_{i,i'} - \h(0)_{i,i'} = \sum_{r=1}^m\sum_{j,j'=1}^pa_{rj}a_{rj'}\inner{\hat{\x}_i^{(j)}}{\hat{\x}_{i'}^{(j')}}z_{r,i,i'}^{(j,j')},
\end{align*}
with
\begin{align*}
    z_{r,i,i'}^{(j,j')} = \indy{\hat{\x}_i^{(j)};\w_{r,t}}\indy{\hat{\x}_{i'}^{(j')};\w_{r,t}} - \indy{\hat{\x}_i^{(j)};\w_{r,0}}\indy{\hat{\x}_{i'}^{(j')};\w_{r,0}}.
\end{align*}
We observe that $|z_{r,i,i'}^{(j,j')}|$ only if $A_{ir}^{(j)}\vee A_{i'r}^{(j')}$. Therefore:
\begin{align*}
    \E_{\mathbf{W}}\left[z_{r,i,i'}^{(j,j')}\right]\leq \mathbb{P}\paren{A_{ir}^{(j)}} + \mathbb{P}\paren{A_{i'r}^{(j')}}\leq \frac{4R}{\kappa\sqrt{2\pi}}.
\end{align*}
For the case $j = j'$, we first notice that
\begin{align*}
    \E_{\mathbf{W}}\left[\paren{z_{r,i,i'}^{(j,j')} - \E_{\mathbf{W}}\left[z_{r,i,i'}^{(j,j')}\right]}^2\right] \leq \E_{\mathbf{W}}\left[z_{r,i,i'}^{(j,j')2}\right] \leq \frac{4R}{\kappa\sqrt{2\pi}}.
\end{align*}
Thus, applying Berstein Inequality to the case $j = j'$ we have that:
\begin{align*}
    \mathbb{P}\paren{\sum_{r=1}^mz_{r,i,i'}^{(j,j)}\geq m\paren{\E_{\mathbf{W}}\left[z_{r,i,i'}^{(j,j')}\right] + t}} \leq \exp\paren{-\frac{\kappa mt^2\sqrt{2\pi}}{8\paren{1 + \frac{t}{3}}R}}.
\end{align*}
For the case $j\neq j'$, we notice that:
\begin{align*}
    \E_{\W,\mathbf{a}}\left[a_{rj}a_{rj'}z_{r,i,i'}^{(j,j')}\right] = 0.
\end{align*}
Moreover:
\begin{align*}
    \left|a_{rj}a_{rj'}z_{r,i,i'}^{(j,j')}\right|& \leq \frac{4R}{p^2m\kappa\sqrt{2\pi}},\\
    \E_{\W,\mathbf{a}}\left[\paren{a_{rj}a_{rj'}z_{r,i,i'}^{(j,j')}}^2\right] & =\frac{1}{p^4m^2}\E_{\W}\left[z_{r,i,i'}^{(j,j')2}\right] \leq\frac{4R}{p^4m^2\kappa\sqrt{2\pi}}.
\end{align*}
Applying the Berstein Inequality to the case $j\neq j'$, we have that:
\begin{align*}
    \mathbb{P}\paren{\sum_{r=1}^ma_{rj}a_{rj'}z_{r,i,i'}^{(j,j')}\geq \frac{t}{p^2}} \leq \exp\paren{-\frac{m\kappa t^2\sqrt{2\pi}}{8\paren{1 + \frac{t}{3}}R}}.
\end{align*}
Combining both cases, we have that with probability at least $1 - p^2\exp\paren{-\frac{m\kappa t^2\sqrt{2\pi}}{8\paren{1 + \frac{t}{3}}R}}$, it holds that:
\begin{align*}
    \left|\h(t)_{i,i'} - \h(0)_{i,i'}\right|\leq p^{-1}\E_{\mathbf{W}}\left[z_{r,i,i'}^{(j,j')}\right] + t \leq \frac{2R}{p\kappa} + t^2.
\end{align*}
Choose $t = \kappa^{-1}R$. Then as long as $m = \frac{\log \frac{np}{\delta}}{R}$, it holds that with probability at least $1 - O(\delta)$:
\begin{align*}
    \left|\h(t)_{i,i'} - \h(0)_{i,i'}\right|\leq 3\kappa^{-1}R.
\end{align*}
This implies that:
\begin{align*}
    \norm{\h(t) - \h(0)}_2\leq\norm{\h(t)-\h(0)}_F \leq 3n\kappa^{-1}R.
\end{align*}
Thus, $\norm{\h(t) - \h(0)}_2\leq \frac{\lambda_0}{4}$ as long as $R = O\paren{\frac{\kappa \lambda_0}{n}}$. This shows that $\lambda_{\min}\paren{\h(t)}\geq \frac{\lambda_0}{2}$ for all $t$ with probability at least $1 - O(\delta)$.

\subsection{Surrogate Gradient Bound}
As we see in previous section, the \texttt{AsyncDrop} scheme can be written as 
\begin{align*}
    \w_{r,t+1} = \w_{r,t} - \eta\mathbf{g}_{r,t-\delta_t};\quad \mathbf{g}_{r,t} = \sum_{i=1}^n\sum_{j=1}^pa_{rj}\paren{\tilde{u}_{r,t}^{(i)} - N_{r,t}^\perp y_i}\hat{\x}_i^{(j)}\indy{\hat{\x}_i^{(j)};\w_{r,t}}
\end{align*}
with $\tilde{u}_{r,t}^{(i)}$ defined as
\begin{align*}
    \tilde{u}_{r,t}^{(i)} = \frac{N_{r,t}^\perp}{N_{r,t}}\sum_{s=1}^Sm_{r,t}^{(s)}\hat{u}_t^{(s, i)} = \sum_{r'=1}^m\sum_{j=1}^p\underbrace{\paren{\frac{N_{r,t}^\perp}{N_{r,t}}\sum_{s=1}^Sm_{r,t}^{(s)}m_{r',t}^{(s)}}}_{\nu_{r,r',t}}a_{rj}\sigma\paren{\inner{\hat{\x}_i^{(j)}}{\w_{r,t}}}
\end{align*}
The mixing of the surrogate function $\tilde{u}_{r,t}^{(i)}$ can be bounded by
\begin{align*}
    \E_{\mathbf{M}_t}\left[\tilde{u}_{r,t}^{(i)}\right] &= \sum_{s=1}^S\sum_{r'=1}^m\sum_{j=1}^p\E_{\mathbf{M}_t}\left[m_{r,t}^{(s)}m_{r,t'}^{(s)}\cdot\frac{N_{r,t}^\perp}{N_{r,t}}\right]a_{r'j}\sigma\paren{\inner{\hat{\x}^{(j)}}{\w_{r,t'}}}\\
    & = \xi\theta\sum_{r'=1}^m\sum_{j=1}^pa_{rj}\sigma\paren{\inner{\hat{\x}^{(j)}}{\w_{r,t}}} + (1-\xi)\theta\sum_{j=1}^pa_{rj}\sigma\paren{\inner{\hat{\x}^{(j)}_i}{\w_{r,t}}}\\
    & = \theta u_t^{(i)} + (1-\xi)\theta\underbrace{\sum_{j=1}^pa_{rj}\sigma\paren{\inner{\hat{\x}^{(j)}_i}{\w_{r,t}}}}_{\hat{\epsilon}_{r,t}^{(i)}}
\end{align*}
Therefore,
\begin{align*}
    \E_{\mathbf{M}_t}\left[\mathbf{g}_{r,t}\right] & = \sum_{i=1}^n\sum_{j=1}^na_{rj}\E_{\mathbf{M}_t}\left[\tilde{u}_t^{(i)} - N_{r,t}^\perp y_i\right]\hat{\x}_i^{(j)}\indy{\hat{\x}_i^{(j)};\w_{r,t}}\\
    & = \xi^{-1}\theta\nabla_{\w_r}\mathcal{L}\paren{\W_t} + (1-\xi)\theta\underbrace{\sum_{i=1}^n\sum_{j'=1}^pa_{rj'}\hat{\epsilon}_{r,t}^{(i)}\hat{\x}_i^{(j)}\indy{\hat{\x}_i^{(j)};\w_{r,t}}}_{\boldsymbol{\epsilon}_{r,t}}
\end{align*}
Now, we have
\begin{align*}
    \left|\hat{\epsilon}_{r,t}^{(i)}\right|\leq \frac{1}{\sqrt{m}}\norm{\w_{r,t}}_2;\quad \norm{\boldsymbol{\epsilon}_{r,t}}_2 \leq \frac{n}{\sqrt{m}}\left|\hat{\epsilon}_{r,t}^{(i)}\right|\leq \frac{n}{m}\norm{\w_{r,t}}_2
\end{align*}
Moreover, we would like to investigate the norm and norm squared of the gradient. In particular, we first notice that, under the case of $N_{t,r}^\perp = 1$, we have
\begin{align*}
    \mathbf{g}_{r,t} = \xi^{-1}\nabla_{\w_r}\mathcal{L}\paren{\W_t} + \sum_{i=1}^n\sum_{j=1}^pa_{rj}\paren{\tilde{u}_{r,t}^{(i)} - u_{t}^{(i)}}\hat{\x}_i^{(j)}\indy{\hat{\x}_i^{(j)};\w_{r,t}}
\end{align*}
Thus, we are interested in $\norm{\tilde{\U}_{r,t} - \U_t}_2$. Following from previous work \cite{Liao2021MaskedNTK} (lemma 19, 20, and 21), we have that
\begin{align*}
    \E_{\mathbf{M}_t}\left[\nu_{r,r',t}\mid N_{r,t}^\perp = 1\right]\begin{cases}
    \xi & \text{ if }r\neq r'\\
    1 & \text{ if }r = r'
    \end{cases}
    \quad\quad 
    \text{Var}\paren{\nu_{r,r',t}\mid N_{r,t}^\perp=1} = \begin{cases}
    \frac{\theta-\xi^2}{S} & \text{ if }r\neq r'\\
    0 & \text{ if }r = r'
    \end{cases}
\end{align*}
Therefore
\begin{align*}
    \E_{\mathbf{M}_t} & \left[\norm{\tilde{\U}_{r,t} - \U_t}_2^2\mid N_{r,t}^\perp=1\right]\\ & = \sum_{i=1}^n\E_{\mathbf{M}_t}\left[\paren{\sum_{r'=1}^m\sum_{j=1}^p\paren{\nu_{r,r',t}-\xi}a_{rj}\sigma\paren{\inner{\hat{\x}_i^{(j)}}{\w_{r',t}}}}^2\mid N_{r,t}^\perp=1\right]\\
    & = \sum_{i=1}^n\sum_{r'=1}^m\E_{\mathbf{M}_t}\left[\paren{\nu_{r,r',t}-\xi}^2\mid N_{r,t}^\perp=1\right]\paren{\sum_{j=1}^pa_{rj}\sigma\paren{\inner{\hat{\x}_i^{(j)}}{\w_{r',t}}}}^2\\
    & \leq \frac{1}{mp}\sum_{i=1}^n\sum_{r'=1}^m\E_{\mathbf{M}_t}\left[\paren{\nu_{r,r',t}-\xi}^2\mid N_{r,t}^\perp=1\right]\sum_{j=1}^p\sigma\paren{\inner{\hat{\x}_i^{(j)}}{\w_{r',t}}}^2\\
    & \leq \frac{1}{mp}\sum_{i=1}^n\sum_{r'\neq r}\text{Var}\paren{\nu_{r,r',t}\mid N_{r,t}^\perp=1}\sum_{j=1}^p\sigma\paren{\inner{\hat{\x}_i^{(j)}}{\w_{r',t}}}^2 + \\
    &\quad\quad\quad \frac{1}{mp}\sum_{i=1}^n\E_{\mathbf{M}_t}\left[\paren{\nu_{r,r,t}-\xi}^2\mid N_{r,t}^\perp=1\right]\sum_{j=1}^p\sigma\paren{\inner{\hat{\x}_i^{(j)}}{\w_{r',t}}}^2\\
    & \leq \frac{1}{mpS}(\theta-\xi)\sum_{r'\neq r}\sum_{i=1}^n\sum_{j=1}^p\inner{\hat{\x}_i^{(j)}}{\w_{r',t}}^2 + \frac{1}{mp}(\theta-\xi^2)\sum_{i=1}^n\sum_{j=1}^p\inner{\hat{\x}_i^{(j)}}{\w_{r,t}}^2
\end{align*}
With high probabilty it holds that
\begin{align*}
    \sum_{r=1}^m\sum_{i=1}^n\sum_{j=1}^p\inner{\hat{\x}_i^{(j)}}{\w_{r,0}}^2 \leq 2mnp\kappa^2 - mnpR^2
\end{align*}
Thus, with sufficiently large $m$, the second term is always smaller than the first term, and we have
\begin{align*}
    \E_{\mathbf{M}_t}\left[\norm{\tilde{\U}_{r,t} - \U_t}_2^2\mid N_{r,t}^\perp=1\right] \leq 8n\kappa^2(\theta-\xi^2)S^{-1}
\end{align*}
Now, we can compute that
\begin{align*}
    \E_{\mathbf{M}_t}\left[\norm{\mathbf{g}_{r,t}}_2\mid N_{r,t}^\perp=1\right] & =  \E_{\mathbf{M}_t}\left[\norm{\sum_{i=1}^n\sum_{j=1}^pa_{rj}\paren{\tilde{u}_{r,t}^{(i)} - u_{t}^{(i)}}\hat{\x}_i^{(j)}\indy{\hat{\x}_i^{(j)};\w_{r,t}}}\mid N_{r,t}=1\right] + \\
    &\quad\quad\quad \xi^{-1}\theta\norm{\nabla_{\w_r}\mathcal{L}\paren{\W_t}}_2\\
    & = \sqrt{\frac{n}{m}}\E_{\mathbf{M}_t}\left[\norm{\tilde{\U}_{r,t} - \U_t}_2\mid N_{r,t}=1\right] + \theta\sqrt{\frac{n}{m}}\norm{\U_t - \y}_2\\
    & \leq n\kappa\sqrt{\frac{\theta-\xi^2}{mS}} + \theta\sqrt{\frac{n}{m}}\norm{\U_t - \y}_2
\end{align*}
And we know that $\mathbf{g}_{r,t} = 0$ when $N_{r,t}^\perp=0$. Therefore,
\begin{align*}
    \E_{\mathbf{M}_t}\left[\norm{\mathbf{g}_{r,t}}_2\right]\leq \theta n\kappa\sqrt{\frac{\theta-\xi^2}{mS}} + \theta^2\sqrt{\frac{n}{m}}\norm{\U_t - \y}_2
\end{align*}
Similarly
\begin{align*}
    \E_{\mathbf{M}_t}\left[\norm{\mathbf{g}_{r,t}}_2^2\mid N_{r,t}^\perp=1\right] & = \frac{2n}{m}\E_{\mathbf{M}_t}\left[\norm{\tilde{\U}_{r,t} - \U_t}_2^2\mid N_{r,t}=1\right] + \frac{2\theta^2n}{m}\norm{\U_t - \y}_2^2\\
    & \leq \frac{16n^2\kappa^2\paren{\theta-\xi^2}}{mS} + \frac{2\theta^2n}{m}\norm{\U_t - \y}_2^2
\end{align*}
and thus
\begin{align*}
    \E_{\mathbf{M}_t}\left[\norm{\mathbf{g}_{r,t}}_2^2\right]\leq \frac{16\theta n^2\kappa^2\paren{\theta-\xi^2}}{mS} + \frac{2\theta^3n}{m}\norm{\U_t - \y}_2^2
\end{align*}

\subsection{Step-wise Convergence}Consider
\begin{align*}
    \E_{\mathbf{M}_t}\left[\norm{\U_{t+1} - \y}_2^2\right] & = \norm{\U_t - \y}_2^2 - 2\inner{\U_t- \U_{t+1}}{\U_t - \y} + \norm{\U_t - \U_{t+1}}_2^2\\
    & = \norm{\U_t - \y}_2^2 - 2\inner{\mathbf{I}_{1,t} + \mathbf{I}_{2,t}}{\U_t - \y} + \norm{\U_t - \U_{t+1}}_2^2\\
    & \leq \norm{\U_t - \y}_2^2 - 2\inner{\mathbf{I}_{1,t}}{\U_t - \y} + 2\norm{\mathbf{I}_{2,t}}_2\norm{\U_t - \y}_2 + \norm{\U_t - \U_{t+1}}_2^2
\end{align*}
Let $P_{ij} = \left\{r\in[m]:\neg A_{ir}^{(j)}\right\}$. Here $\mathbf{I}_{1,t}$ and $\mathbf{I}_{2,t}$ are characterized as in previous work.
\begin{align*}
I_{1,t}^{(i)} & = \sum_{j=1}^p\sum_{r\in P_{ij}}a_{rj}\paren{\sigma\paren{\inner{\xij}{\w_{r,t}}} - \sigma\paren{\inner{\xij}{\w_{r,t+1}}}}\\
    I_{2,t}^{(i)} & = \sum_{j=1}^p\sum_{r\in P_{ij}^\perp}a_{rj}\paren{\sigma\paren{\inner{\xij}{\w_{r,t}}} - \sigma\paren{\inner{\xij}{\w_{r,t+1}}}}
\end{align*}

We first bound the magnitude of $\mathbf{I}_{2,t}$
\begin{align*}
    \left|I_{2,t}^{(i)}\right| & = \frac{1}{p\sqrt{m}}\sum_{j=1}^p\sum_{r\in P_{ij}^\perp}\left|\paren{\sigma\paren{\inner{\hat{\x}_i^{(j)}}{\w_{r,t}}}-\sigma\paren{\inner{\hat{\x}_i^{(j)}}{\w_{r,t+1}}}}\right|\\
    & \leq \frac{1}{p\sqrt{m}}\sum_{j=1}^p\sum_{r\in P_{ij}}\norm{\w_{r,t} - \w_{r,t+1}}_2\\
    & \leq \frac{\eta}{p\sqrt{m}} \sum_{j=1}^p\sum_{r\in P_{ij}}\norm{\mathbf{g}_{r,t-\delta_t}}_2\\
    & \leq \frac{\eta}{\sqrt{m}}\cdot 3m\kappa^{-1}R\cdot \paren{n\kappa\sqrt{\frac{\theta-\xi^2}{mS}} + \theta\sqrt{\frac{n}{m}}\norm{\U_{t-\delta_t} - \y}_2}\\
    & = 3\kappa^{-1}\theta\eta\sqrt{n}R\norm{\U_{t-\delta_t} -\y}_2 + 3\eta nR\sqrt{\frac{\theta-\xi^2}{S}}
\end{align*}
Thus
\begin{align*}
    \E_{\mathbf{M}_t}\left[\norm{\mathbf{I}_{2,t}}_2\right] & \leq \E_{\mathbf{M}_t}\left[\sqrt{n}\max_{i\in[n]}\left|I_{2,t}^{(i)}\right|\right]\\
    & \leq \frac{\eta}{p} \sqrt{\frac{n}{m}}\max_{i\in[n]}\sum_{j=1}^p\sum_{r\in P_{ij}}\E_{\mathbf{M}_t}\left[\norm{\mathbf{g}_{r,t-\delta_t}}_2\right]\\
    & \leq \eta\sqrt{\frac{n}{m}}\cdot 3m\kappa^{-1}R\cdot \paren{n\kappa\theta\sqrt{\frac{\theta-\xi^2}{mS}} + \theta^2\sqrt{\frac{n}{m}}\norm{\U_{t-\delta_t} - \y}_2}\\
    & = 3\kappa^{-1}\theta^2\eta nR\norm{\U_{t-\delta_t} -\y}_2 + 3\theta\eta n^{\frac{3}{2}}R\sqrt{\frac{\theta-\xi^2}{S}}
\end{align*}
Therefore, given assumption $\norm{\U_{t-\delta_t} -\y}_2 \geq \norm{\U_{t} -\y}_2$
\begin{align*}
    \E_{\mathbf{M}_t}\left[\norm{\mathbf{I}_{2,t}}_2\norm{\U_{t} - \y}_2\right] \leq 6\theta\eta n\kappa^{-1}R\norm{\U_{t-\delta_t} - \y}_2^2 + 3S^{-1}\theta\paren{\theta-\xi^2}\eta n^2\kappa R
\end{align*}
Letting $R = O\paren{\frac{\kappa\lambda_0}{n}}$ gives that
\begin{align*}
    \E_{\mathbf{M}_t}\left[\norm{\mathbf{I}_{2,t}}_2\norm{\U_{t} - \y}_2\right] = O\paren{\theta\eta\lambda_0 }\norm{\U_{t-\delta_t} - \y}_2^2 + O\paren{\theta\eta\lambda_0(\theta-\xi^2)S^{-1}n\kappa^2}
\end{align*}

As in previous work, $I_{1,t}^{(i)}$ can be written as
\begin{align*}
    \E_{\mathbf{M}_t}\left[I_{1,t}^{(i)}\right] & = \xi\sum_{j=1}^p\sum_{r\in P_{ij}}a_{rj}\E_{\mathbf{M}_t}\left[\sigma\paren{\inner{\hat{\x}_i^{(j)}}{\w_{r,t}}}-\sigma\paren{\inner{\hat{\x}_i^{(j)}}{\w_{r,t+1}}}\right]\\
    & = \xi\sum_{j=1}^p\sum_{r\in P_{ij}}a_{rj}\inner{\hat{\x}_i^{(j)}}{\E_{\mathbf{M}_t}\left[\w_{r,t} - \w_{r,t+1}\right]}\indy{\hat{\x}_i^{(j)};\w_{r,t}}\\
    & = \xi\eta\sum_{j=1}^p\sum_{r\in P_{ij}}a_{rj}\inner{\hat{\x}_i^{(j)}}{\E_{\mathbf{M}_t}\left[\mathbf{g}_{r,t-\delta_t}\right]}\indy{\hat{\x}_i^{(j)};\w_{r,t}}\\
    & = \xi\eta\sum_{j=1}^p\sum_{r\in P_{ij}}a_{rj}\inner{\hat{\x}_i^{(j)}}{\E_{\mathbf{M}_t}\left[\mathbf{g}_{r,t-\delta_t}-\mathbf{g}_{r,t}+\mathbf{g}_{r,t}\right]}\indy{\hat{\x}_i^{(j)};\w_{r,t}}\\
    & = \xi\eta\sum_{j=1}^p\sum_{r\in P_{ij}}a_{rj}\inner{\hat{\x}_i^{(j)}}{\E_{\mathbf{M}_t}\left[\mathbf{g}_{r,t}\right]}\indy{\hat{\x}_i^{(j)};\w_{r,t}} + \\ 
    &\quad\quad\quad\xi\eta\sum_{j=1}^p\sum_{r\in P_{ij}}a_{rj}\inner{\hat{\x}_i^{(j)}}{\E_{\mathbf{M}_t}\left[\mathbf{g}_{r,t-\delta_t}-\mathbf{g}_{r,t}\right]}\indy{\hat{\x}_i^{(j)};\w_{r,t}} \\
    & = \eta\theta\sum_{j=1}^p\sum_{r\in P_{ij}}a_{rj}\inner{\hat{\x}_i^{(j)}}{\nabla_{\w_r}\mathcal{L}\paren{\W_t}}\indy{\hat{\x}_i^{(j)};\w_{r,t}} \\
    &\quad\quad\quad + \xi(1-\xi)\theta\eta\sum_{j=1}^p\sum_{r\in P_{ij}}a_{rj}\inner{\hat{\x}_i^{(j)}}{\boldsymbol{\epsilon}_{r,t}}\indy{\hat{\x}_i^{(j)};\w_{r,t}} \\
    & \quad\quad\quad +\eta\theta\sum_{j=1}^p\sum_{r\in P_{ij}}a_{rj}\inner{\hat{\x}_i^{(j)}}{\nabla_{\w_r}\mathcal{L}\paren{\W_{t-\delta_t}}-\nabla_{\w_r}\mathcal{L}\paren{\W_t}}\indy{\hat{\x}_i^{(j)};\w_{r,t}} \\
    & \quad\quad\quad + \xi(1-\xi)\theta\eta\sum_{j=1}^p\sum_{r\in P_{ij}}a_{rj}\inner{\hat{\x}_i^{(j)}}{\boldsymbol{\epsilon}_{r,t-\delta_t}-\boldsymbol{\epsilon}_{r,t}}\indy{\hat{\x}_i^{(j)};\w_{r,t}}
\end{align*}
\begin{align*}
    & = \eta\theta\xi\sum_{i'=1}^n\sum_{j,j'=1}^p\sum_{r\in P_{ij}}\paren{u_t^{(i')} - y_{i'}}a_{rj}a_{rj'}\inner{\hat{\x}_i^{(j)}}{\hat{\x}_{i'}^{(j')}}\indy{\hat{\x}_i^{(j)};\w_{r,t}}\cdot\indy{\hat{\x}_{i'}^{(j')};\w_{r,t}} \\
    &\quad\quad\quad + \xi(1-\xi)\theta\eta\sum_{j=1}^p\sum_{r\in P_{ij}}a_{rj}\inner{\hat{\x}_i^{(j)}}{\boldsymbol{\epsilon}_{r,t}}\indy{\hat{\x}_i^{(j)};\w_{r,t}} \\
    & \quad\quad\quad + \eta\theta\sum_{j=1}^p\sum_{r\in P_{ij}}a_{rj}\inner{\hat{\x}_i^{(j)}}{\nabla_{\w_r}\mathcal{L}\paren{\W_{t-\delta_t}}-\nabla_{\w_r}\mathcal{L}\paren{\W_t}}\indy{\hat{\x}_i^{(j)};\w_{r,t}} \\
    & \quad\quad\quad + \xi(1-\xi)\theta\eta\sum_{j=1}^p\sum_{r\in P_{ij}}a_{rj}\inner{\hat{\x}_i^{(j)}}{\boldsymbol{\epsilon}_{r,t-\delta_t}-\boldsymbol{\epsilon}_{r,t}}\indy{\hat{\x}_i^{(j)};\w_{r,t}}\\
    & = \eta\theta\sum_{i'=1}^n\paren{\h(t)_{ii'} - \h(t)^\perp_{ii'}}\paren{u_t^{(i')}-y_{i'}} \\
    & \quad\quad\quad + \underbrace{\xi(1-\xi)\theta\eta\sum_{j=1}^p\sum_{r\in P_{ij}}a_{rj}\inner{\hat{\x}_i^{(j)}}{\boldsymbol{\epsilon}_{r,t}}\indy{\hat{\x}_i^{(j)};\w_{r,t}}}_{\gamma_{1,i,t}} \\
    & \quad\quad\quad + \underbrace{\eta\theta\sum_{j=1}^p\sum_{r\in P_{ij}}a_{rj}\inner{\hat{\x}_i^{(j)}}{\nabla_{\w_r}\mathcal{L}\paren{\W_{t-\delta_t}}-\nabla_{\w_r}\mathcal{L}\paren{\W_t}}\indy{\hat{\x}_i^{(j)};\w_{r,t}}}_{\gamma_{2,i,t}} \\
    & \quad\quad\quad + \underbrace{\xi(1-\xi)\theta\eta\sum_{j=1}^p\sum_{r\in P_{ij}}a_{rj}\inner{\hat{\x}_i^{(j)}}{\boldsymbol{\epsilon}_{r,t-\delta_t}-\boldsymbol{\epsilon}_{r,t}}\indy{\hat{\x}_i^{(j)};\w_{r,t}}}_{\gamma_{3,i,t}}\\
\end{align*}
Note that: \vspace{-1cm}
\begin{align*}
    \left|\gamma_{1,i,t}\right| & \leq \xi(1-\xi)\theta\eta m^{-\frac{1}{2}}\sum_{r=1}^m\norm{\boldsymbol{\epsilon}_{r,t}} \\ &\leq \xi(1-\xi)\theta\eta nm^{-1}\norm{\W_{t}}_F \\ 
    &\leq O\paren{\xi(1-\xi)\theta\eta n\kappa\sqrt{\frac{d}{m}}}
\end{align*}
Given that: \vspace{-1cm}
\begin{align*}
\norm{\paren{u_{t-\delta_t}^{(i)} - u_{t}^{(i)}}}_2 & = \norm {\xi\sum_{r=1}^m\sum_{j=1}^pa_{rj}\sigma\paren{\inner{\hat{\x}_i^{(j)}}{\w_{r,t-\delta_t}}}-\xi\sum_{r=1}^m\sum_{j=1}^pa_{rj}\sigma\paren{\inner{\hat{\x}_i^{(j)}}{\w_{r,t}}}}_2 \\
& \leq \xi\sum_{r=1}^m\sum_{j=1}^p \norm{a_{rj} \hat{\x}_i^{(j)}}_2 \norm{\w_{r,t-\delta_t}-\w_{r,t}}_2 \leq \xi \frac{1}{\sqrt{m}} R_E
\end{align*}

We show:
\begin{align*}
 \|\nabla_{\w_r}\mathcal{L}\paren{\W_{t-\delta_t}}-&\nabla_{\w_r}\mathcal{L}\paren{\W_t}\|_2 \\
&= \norm{\xi\sum_{i=1}^n\sum_{j=1}^p\paren{u_{t-\delta_t}^{(i)} - y_i}a_{rj}\xij\indy{\xij;\w_{r,t-\delta_t}}-\xi\sum_{i=1}^n\sum_{j=1}^p\paren{u_t^{(i)} - y_i}a_{rj}\xij\indy{\xij;\w_{r,t}}}_2 \\
&\leq   \xi\sum_{i=1}^n\sum_{j=1}^p \norm{a_{rj} \xij}_2 \norm{\paren{u_{t-\delta_t}^{(i)} - y_i}\indy{\xij;\w_{r,t-\delta_t}}-\paren{u_{t}^{(i)} - y_i}\indy{\xij;\w_{r,t}}}_2 \\
&\leq   \xi\sum_{i=1}^n\sum_{j=1}^p \norm{a_{rj} \xij}_2 \left(\norm{\paren{u_{t-\delta_t}^{(i)} - u_{t}^{(i)}}}_2+\norm{\paren{u_{t}^{(i)} - y_i}}_2\right) \\
&\leq   \xi^2 \frac{n}{m} R_E+ \xi \sqrt{\frac{n}{m}} \norm{\U_t - \y}_2
\end{align*}

Thus, 
\begin{align*}
\norm{\gamma_{2,i,t}}_2 & \leq 2\eta\theta n m^{-2}\xi^2 R_E+ \eta\theta\xi n^{\frac{1}{2}}m^{-\frac{3}{2}} \norm{\U_t - \y}_2
\end{align*}

Given that:
\begin{align*}
\norm{\hat{\epsilon}_{r,t-\delta_t}^{(i)} - \hat{\epsilon}_{r,t}^{(i)}}_2 &= \norm{\sum_{j=1}^pa_{rj}\sigma\paren{\inner{\hat{\x}^{(j)}_i}{\w_{r,t-\delta_t}}} -\sum_{j=1}^pa_{rj}\sigma\paren{\inner{\hat{\x}^{(j)}_i}{\w_{r,t}}} }_2 \\
& \leq \sum_{j=1}^p \norm{a_{rj}\hat{\x}^{(j)}_i}_2 (\norm{\w_{r,t-\delta_t}-\w_{r,t}}_2+\norm{\w_{r,t}}_2) \\
& \leq \kappa\sqrt{\frac{2d}{m}}+\frac{1}{\sqrt{m}} R_E
\end{align*}

We show:
\begin{align*}
\norm{\boldsymbol{\epsilon}_{r,t-\delta_t}^{(i)}-\boldsymbol{\epsilon}_{r,t}^{(i)}}_2 & =  \norm{\sum_{i=1}^n\sum_{j'=1}^pa_{rj'}\hat{\epsilon}_{r,t-\delta_t}^{(i)}\hat{\x}_i^{(j)}\indy{\hat{\x}_i^{(j)};\w_{r,t-\delta_t}}-\sum_{i=1}^n\sum_{j'=1}^pa_{rj'}\hat{\epsilon}_{r,t}^{(i)}\hat{\x}_i^{(j)}\indy{\hat{\x}_i^{(j)};\w_{r,t}}}_2 \\
& \leq \sum_{i=1}^n\sum_{j'=1}^p \norm{a_{rj'}\hat{\x}_i^{(j)}}_2 \left(\norm{\paren{\hat{\epsilon}_{r,t-\delta_t}^{(i)} - \hat{\epsilon}_{r,t}^{(i)}}}_2+\norm{\paren{\hat{\epsilon}_{r,t}^{(i)}}}_2\right) \\
& \leq \frac{n}{\sqrt{m}} \left(\kappa\sqrt{\frac{2d}{m}}+\frac{1}{\sqrt{m}} R_E +\frac{1}{\sqrt{m}} \kappa\sqrt{2dm}\right)
\end{align*}

Thus,
\begin{align*}
    \norm{\gamma_{3,i,t}}_2 & \leq \sqrt{2}\xi(1-\xi)\theta\eta n m^{-\frac{3}{2}} \kappa\sqrt{d}+ 2\xi(1-\xi)\theta\eta n m^{-\frac{3}{2}} R_E + \sqrt{2} \xi(1-\xi)\theta\eta n m^{-1} \kappa\sqrt{d})
\end{align*}
This implies that:
\begin{align*}
    \E_{\mathbf{M}_t}\left[\inner{\mathbf{I}_{1,t}}{\U_t - \y}\right] & = \eta\theta\sum_{i,i'=1}^n\paren{u_t^{(i)}-y_{i}}\paren{\h(t)_{ii'} - \h(t)^\perp_{ii'}}\paren{u_t^{(i')}-y_{i'}} \\
    &\quad\quad\quad + \sum_{i=1}^n\gamma_{1,i,t}\paren{u_t^{(i)}-y_i} + \sum_{i=1}^n\gamma_{2,i,t}\paren{u_t^{(i)}-y_i} + \sum_{i=1}^n\gamma_{3,i,t}\paren{u_t^{(i)}-y_i}\\
    & = \eta\theta\inner{\U_t - \y}{\paren{\h(t) - \h(t)^\perp}\paren{\U_t - \y}} \\
    &\quad\quad\quad + \sum_{i=1}^n\gamma_{1,i,t}\paren{u_t^{(i)}-y_i} + \sum_{i=1}^n\gamma_{2,i,t}\paren{u_t^{(i)}-y_i} + \sum_{i=1}^n\gamma_{3,i,t}\paren{u_t^{(i)}-y_i}\\
    & \geq \eta\theta\paren{\lambda_{\min}\paren{\h(t)} - \lambda_{\max}\paren{\h(t)^\perp}}\norm{\U_t - \y}_2^2 \\
    &\quad\quad\quad - \sum_{i=1}^n\left|\gamma_{1,i,t}\right|\cdot\left|u_t^{(i)} - y_i\right|- \sum_{i=1}^n\left|\gamma_{2,i,t}\right|\cdot\left|u_t^{(i)} - y_i\right|- \sum_{i=1}^n\left|\gamma_{3,i,t}\right|\cdot\left|u_t^{(i)} - y_i\right|\\
    & \geq \eta\theta\paren{\lambda_{\min}\paren{\h(t)} - \lambda_{\max}\paren{\h(t)^\perp}}\norm{\U_t - \y}_2^2 \\
    &\quad\quad\quad - \sqrt{n}\max_{i}\left|\gamma_{i,t}\right|\norm{\U_t - \y}_2\\
    & \quad\quad\quad 2\eta\theta - n^{\frac{3}{2}} m^{-2}\xi^2R_E \norm{\U_t - \y}_2 - \eta\theta\xi nm^{-\frac{3}{2}} \norm{\U_t - \y}_2^2 \\
    & \quad\quad\quad - \sqrt{2}\xi(1-\xi)\theta\eta n m^{-\frac{3}{2}} \kappa\sqrt{d}\norm{\U_t - \y}_2 -2\xi(1-\xi)\theta\eta n m^{-\frac{3}{2}} R_E\norm{\U_t - \y}_2 \\
    & \quad\quad\quad\sqrt{2} - \xi(1-\xi)\theta\eta n m^{-1} \kappa\sqrt{d} \norm{\U_t - \y}_2\\
    & \geq \eta\theta\paren{\lambda_{\min}\paren{\h(t)} - \lambda_{\max}\paren{\h(t)^\perp} - O\paren{\lambda_0}-nm^{-\frac{3}{2}}}\norm{\U_t - \y}_2^2 \\
    &\quad\quad\quad - O\left(\frac{\xi^2(1-\xi)^2\theta\eta n^3\kappa^2d}{m\lambda_0} + \frac{\eta^2\theta^2 n^{3} \xi^4R_E^2}{m^{4}\lambda_0} + \frac{\xi^2(1-\xi)^2\theta^2\eta^2 n^2  \kappa^2 d}{m^3\lambda_0 }+ \right.
\end{align*}
\begin{align*}
    &\left. \quad\quad\quad \frac{\xi^2(1-\xi)^2\theta^2\eta^2 n^2 R_E^2}{m^3\lambda_0} + \frac{\xi^2(1-\xi)^2\theta^2\eta^2 n^2  \kappa^2 d}{m^2\lambda_0 }\right) \\
\end{align*}
For $\h(t)^\perp$ we have that
\begin{align*}
    \lambda_{\max}\paren{\h(t)^\perp}^2 & \leq \norm{\h(t)^\perp}_F^2\\
    & \leq \sum_{i,i'=1}^n\paren{\sum_{j,j'=1}^p\sum_{r\in P_{ij}}a_{rj}a_{rj'}\inner{\xij}{\hat{\x}_{i,i'}^{(j,j')}}\indy{\xij;\w_{r,t}}\indy{\hat{\x}_{i,i'}^{(j,j')};\w_{r,t}}}^2\\
    & \leq \frac{n^2}{m^2}\paren{\max_{ij}\left|P_{ij}\right|}^2\\
    & \leq n^2\kappa^{-2} R^2
\end{align*}
Choosing $R = O\paren{\frac{\kappa\lambda_0}{n}}$ gives
\begin{align*}
    \lambda_{\max}\paren{\h(t)^\perp} \leq O\paren{\lambda_0}
\end{align*}

We require $m \geq \Omega\paren{\frac{n}{\lambda_0}}$, such that
\begin{align*}
    n m^{-\frac{3}{2}} \leq O\paren{\lambda_0}
\end{align*}

Thus, 
Plugging in $\lambda_{\min}\paren{\h(t)} \geq \frac{\lambda_0}{2}$, we have
\begin{align*}
    \E_{\mathbf{M}_t}\left[\inner{\mathbf{I}_{1,t}}{\U_t - \y}\right] & \geq \eta\theta\lambda_0\paren{\frac{1}{2} - O\paren{1}}\norm{\U_t - \y}_2^2 \\ 
    &\quad\quad\quad - O\left(\frac{\xi^2(1-\xi)^2\theta\eta n^3\kappa^2d}{m\lambda_0} + \frac{\eta^2\theta^2 n \kappa ^2 \lambda_0 \xi^4 E^2}{m^{4}} + \frac{\xi^2(1-\xi)^2\theta^2\eta^2 n^2  \kappa^2 d}{m^3\lambda_0 }+ \right. \\
    &\left. \quad\quad\quad\quad\quad \frac{\xi^2(1-\xi)^2\theta^2\eta^2 \kappa^2 \lambda_0 E^2}{m^3} + \frac{\xi^2(1-\xi)^2\theta^2\eta^2 n^2  \kappa^2 d}{m^2\lambda_0 }\right).
\end{align*}
Lastly, we analyze the last term in the quadratic expansion
\begin{align*}
    \E_{\mathbf{M}_t}\left[\norm{\U_t - \U_{t+1}}_2^2\right] & = \sum_{i=1}^n\E_{\mathbf{M}_t}\left[\paren{u_t^{(i)} - u_{t+1}^{(i)}}^2\right]\\
    & \leq p^{-1}\sum_{i=1}^n\sum_{j=1}^p\sum_{r=1}^m\E_{\mathbf{M}_t}\left[\paren{\sigma\paren{\inner{\xij}{\w_{r,t}}} - \sigma\paren{\inner{\xij}{\w_{r,t+1}}}}^2\right]\\
    & \leq p^{-1}\eta^2\sum_{i=1}^n\sum_{j=1}^p\sum_{r=1}^m\E_{\mathbf{M}_t}\left[\norm{\mathbf{g}_{r,t-\delta_t}}_2^2\right]\\
    & \leq O\paren{\theta^3\eta^2n^2}\norm{\U_{t-\delta_t} - \y}_2^2 + O\paren{\theta\eta^2n^2\kappa^2\paren{\theta-\xi^2}S^{-1}}
\end{align*}
Letting $\eta = O\paren{\frac{\lambda_0}{n^2}}$ gives
\begin{align*}
    \E_{\mathbf{M}_t}\left[\norm{\U_t - \U_{t+1}}_2^2\right] \leq O\paren{\theta\eta\lambda_0}\norm{\U_{t-\delta_t} - \y}_2^2 + O\paren{\theta\eta\lambda_0\kappa^2\paren{\theta-\xi^2}S^{-1}}
\end{align*}

Putting all three terms together and as $\norm{\U_{t-\delta_t} - \y}_2^2 \leq \norm{\U_{t-E} - \y}_2^2$, we have that

\begin{align*}
    \E_{\mathbf{M}_t}\left[\norm{\U_{t+1} - \y}_2^2\right] & \leq \norm{\U_t -  \y}_2^2-\eta\theta\lambda_0\paren{1 - O\paren{1}}\norm{\U_t - \y}_2^2 + O\paren{\theta\eta\lambda_0}\norm{\U_{t-E} - \y}_2^2 \\
    &\quad\quad + O\left(\frac{\xi^2(1-\xi)^2\theta\eta n^3\kappa^2d}{m\lambda_0} + \frac{\eta^2\theta^2 n \kappa ^2 \lambda_0 \xi^4 E^2}{m^{4}} + \frac{\xi^2(1-\xi)^2\theta^2\eta^2 n^2  \kappa^2 d}{m^3\lambda_0 }+ \right. \\
    &\left. \quad\quad\quad\quad\quad \frac{\xi^2(1-\xi)^2\theta^2\eta^2 \kappa^2 \lambda_0 E^2}{m^3} + \frac{\xi^2(1-\xi)^2\theta^2\eta^2 n^2  \kappa^2 d}{m^2\lambda_0 } +\frac{n\kappa^2\paren{\theta-\xi^2}}{S}\right) \\
    & \leq \norm{\U_t -  \y}_2^2-\eta\theta\lambda_0\paren{1 - O\paren{1}}\norm{\U_t - \y}_2^2 + O\paren{\theta\eta\lambda_0}(2\norm{\U_{t-E} - \U_t}_2^2 + 2\norm{\U_t-\y}_2^2)\\
    &\quad\quad + O\left(\frac{\xi^2(1-\xi)^2\theta\eta n^3\kappa^2d}{m\lambda_0} + \frac{\eta^2\theta^2 n \kappa ^2 \lambda_0 \xi^4 E^2}{m^{4}} + \frac{\xi^2(1-\xi)^2\theta^2\eta^2 n^2  \kappa^2 d}{m^3\lambda_0 }+ \right. \\
    &\left. \quad\quad\quad\quad\quad \frac{\xi^2(1-\xi)^2\theta^2\eta^2 \kappa^2 \lambda_0 E^2}{m^3} + \frac{\xi^2(1-\xi)^2\theta^2\eta^2 n^2  \kappa^2 d}{m^2\lambda_0 }+\frac{n\kappa^2\paren{\theta-\xi^2}}{S}\right) \\
    & \leq \norm{\U_t -  \y}_2^2-\eta\theta\lambda_0\paren{1 - O\paren{1}}\norm{\U_t - \y}_2^2 + O\paren{\theta\eta\lambda_0}(8\xi^2 R_E^2 + 2\norm{\U_t-\y}_2^2)\\
    &\quad\quad + O\left(  \frac{\xi^2(1-\xi)^2\theta\eta n^3\kappa^2d}{m\lambda_0} + \frac{\eta^2\theta^2 n \kappa ^2 \lambda_0 \xi^4 E^2}{m^{4}} + \frac{\xi^2(1-\xi)^2\theta^2\eta^2 n^2  \kappa^2 d}{m^3\lambda_0 }+ \right. \\
    &\left. \quad\quad\quad\quad\quad \frac{\xi^2(1-\xi)^2\theta^2\eta^2 \kappa^2 \lambda_0 E^2}{m^3} + \frac{\xi^2(1-\xi)^2\theta^2\eta^2 n^2  \kappa^2 d}{m^2\lambda_0 }+\frac{n\kappa^2\paren{\theta-\xi^2}}{S}\right) \\
\end{align*}

Letting $O\paren{\theta\eta\lambda_0}=\frac{1}{4}\theta\eta\lambda_0$,

\begin{align*}
 \E_{\mathbf{M}_t}\left[\norm{\U_{t+1} - \y}_2^2\right] & \leq \paren{1 - \frac{\theta\eta\lambda_0}{4}}\norm{\U_t- \y}_2^2 + \\
    & \quad\quad\quad O\left(\frac{\theta\eta\lambda_0^3 \xi^2 \kappa^2 E^2}{n^2}+\frac{\xi^2(1-\xi)^2\theta\eta n^3\kappa^2d}{m\lambda_0} + \frac{\eta^2\theta^2 n \kappa ^2 \lambda_0 \xi^4 E^2}{m^{4}} + \frac{\xi^2(1-\xi)^2\theta^2\eta^2 n^2  \kappa^2 d}{m^3\lambda_0 }+ \right. \\
    &\left. \quad\quad\quad\quad \frac{\xi^2(1-\xi)^2\theta^2\eta^2 \kappa^2 \lambda_0 E^2 }{m^3} + \frac{\xi^2(1-\xi)^2\theta^2\eta^2 n^2  \kappa^2 d}{m^2\lambda_0 }+\frac{n\kappa^2\paren{\theta-\xi^2}}{S}\right) \\
\end{align*}

Thus, 

\begin{align*}
\E_{\mathbf{M}_t}\left[\norm{\U_{t+1} - \y}_2^2\right] & \leq \paren{1 - \frac{\theta\eta\lambda_0}{4}}^{t}\norm{\U_0- \y}_2^2 \\
    & \quad\quad\quad + O\left(\frac{\theta\eta\lambda_0^3 \xi^2 \kappa^2 E^2}{n^2}+\frac{\xi^2(1-\xi)^2\theta\eta n^3\kappa^2d}{m\lambda_0} + \frac{\eta^2\theta^2 n \kappa ^2 \lambda_0 \xi^4 E^2}{m^{4}} + \frac{\xi^2(1-\xi)^2\theta^2\eta^2 n^2  \kappa^2 d}{m^3\lambda_0 }+ \right. \\
    &\left. \quad\quad\quad\quad \frac{\xi^2(1-\xi)^2\theta^2\eta^2 \kappa^2 \lambda_0 E^2}{m^3} + \frac{\xi^2(1-\xi)^2\theta^2\eta^2 n^2  \kappa^2 d}{m^2\lambda_0 }+\frac{n\kappa^2\paren{\theta-\xi^2}}{S}\right) \\
\end{align*}

\subsection{Bounding Weight Perturbation}
Next we show that $\norm{\w_{r,t} - \w_{r,0}}_2\leq R$ under sufficient over-parameterization. To start, we notice that
\begin{align*}
    \E_{[\mathbf{M}_{t-1}]}[\|\w_{r,t}&-\w_{r,0}\|_2] \leq \sum_{t'=0}^{t-1}\E_{[\mathbf{M}_{t'}]}\left[\norm{\w_{r,t'+1} - \w_{r,t'}}_2\right]\\
    & \leq \eta\sum_{t'=0}^{t-1}\E_{[\mathbf{M}_{t'}]}\left[\norm{\mathbf{g}_{r,t'-\delta_{t'}}}_2\right]\\
    & \leq \eta\sum_{t'=0}^{t-1}\paren{\theta^2\sqrt{\frac{n}{m}}\E_{[\mathbf{M}_{t'-1}]}\left[\norm{\U_{t'-\delta_{t'}} - \y}_2\right] + \theta n\kappa\sqrt{\frac{\theta-\xi^2}{mS}}}\\
    & \leq \eta \theta^2\sqrt{\frac{n}{m}}\sum_{t'=0}^{t-1}\E_{[\mathbf{M}_{t'-1}]}\left[\norm{\U_{t'-\delta_{t'}} - \y}_2\right] + \eta t \theta n\kappa\sqrt{\frac{\theta-\xi^2}{mS}}\\
    & \leq \eta\theta \sqrt{\frac{n}{m}}\norm{\U_0 - \y}_2\sum_{t'=0}^{t-1}\paren{1 - \frac{\eta\theta\lambda_0}{4}}^{t'} + \eta T \theta  n\kappa\sqrt{\frac{\theta-\xi^2}{mS}} \\
    &\quad\quad\quad + \eta\theta T\cdot O\left(\frac{\theta\eta\lambda_0^3 \xi^2 \kappa^2}{n^2}+\frac{\xi^2(1-\xi)^2\theta\eta n^3\kappa^2d}{m\lambda_0} + \frac{\eta^2\theta^2 n \kappa ^2 \lambda_0 \xi^4}{m^{4}} + \frac{\xi^2(1-\xi)^2\theta^2\eta^2 n^2  \kappa^2 d}{m^3\lambda_0 }+ \right. \\
    &\left. \quad\quad\quad\quad\quad\quad\quad \frac{\xi^2(1-\xi)^2\theta^2\eta^2 \kappa^2 \lambda_0}{m^3} + \frac{\xi^2(1-\xi)^2\theta^2\eta^2 n^2  \kappa^2 d}{m^2\lambda_0 }\right) \\
    & \leq \eta\theta \sqrt{\frac{n}{m}} \frac{1-\paren{1 - \frac{\eta\theta\lambda_0}{4}}^{t}}{\frac{\eta\theta\lambda_0}{4}}\norm{\U_0 - \y}_2  \\
    &\quad\quad\quad + \eta\theta T\cdot O\left(\frac{\theta\eta\lambda_0^3 \xi^2 \kappa^2}{n^2}+\frac{\xi^2(1-\xi)^2\theta\eta n^3\kappa^2d}{m\lambda_0} + \frac{\eta^2\theta^2 n \kappa ^2 \lambda_0 \xi^4}{m^{4}} + \frac{\xi^2(1-\xi)^2\theta^2\eta^2 n^2  \kappa^2 d}{m^3\lambda_0 }+ \right. \\
    &\left. \quad\quad\quad\quad\quad\quad\quad \frac{\xi^2(1-\xi)^2\theta^2\eta^2 \kappa^2 \lambda_0}{m^3} + \frac{\xi^2(1-\xi)^2\theta^2\eta^2 n^2  \kappa^2 d}{m^2\lambda_0 }\right) \\
     & \leq \eta\theta \sqrt{\frac{n}{m}} \frac{1-\paren{1 - \frac{\eta\theta\lambda_0t}{4}}}{\frac{\eta\theta\lambda_0}{4}}\norm{\U_0 - \y}_2  \\
    &\quad\quad\quad + \eta\theta T\cdot O\left(\frac{\theta\eta\lambda_0^3 \xi^2 \kappa^2}{n^2}+\frac{\xi^2(1-\xi)^2\theta\eta n^3\kappa^2d}{m\lambda_0} + \frac{\eta^2\theta^2 n \kappa ^2 \lambda_0 \xi^4}{m^{4}} + \frac{\xi^2(1-\xi)^2\theta^2\eta^2 n^2  \kappa^2 d}{m^3\lambda_0 }+ \right. \\
    &\left. \quad\quad\quad\quad\quad\quad\quad \frac{\xi^2(1-\xi)^2\theta^2\eta^2 \kappa^2 \lambda_0}{m^3} + \frac{\xi^2(1-\xi)^2\theta^2\eta^2 n^2  \kappa^2 d}{m^2\lambda_0 }\right) \\
    & \leq O \paren{\eta\theta \lambda_0 t \sqrt{\frac{n}{m}}} \norm{\U_0 - \y}_2  \\
    &\quad\quad\quad + \eta\theta T\cdot O\left(\frac{\theta\eta\lambda_0^3 \xi^2 \kappa^2}{n^2}+\frac{\xi^2(1-\xi)^2\theta\eta n^3\kappa^2d}{m\lambda_0} + \frac{\eta^2\theta^2 n \kappa ^2 \lambda_0 \xi^4}{m^{4}} + \frac{\xi^2(1-\xi)^2\theta^2\eta^2 n^2  \kappa^2 d}{m^3\lambda_0 }+ \right. \\
    &\left. \quad\quad\quad\quad\quad\quad\quad \frac{\xi^2(1-\xi)^2\theta^2\eta^2 \kappa^2 \lambda_0}{m^3} + \frac{\xi^2(1-\xi)^2\theta^2\eta^2 n^2  \kappa^2 d}{m^2\lambda_0 }\right) \\
\end{align*}
where the last inequality follows from the geometric sum and $\beta \leq O\paren{p^{-1}}$. Using the initialization scale, we have that
\begin{align*}
    \E_{\W,\mathbf{a},[\mathbf{M}_t]}\left[\norm{\w_{r,t}-\w_{r,0}}_2\right] & \leq O \paren{\eta\theta \lambda_0 t nm^{-\frac{1}{2}}} + \eta\theta  T\cdot O\left(\frac{\theta\eta\lambda_0^3 \xi^2 \kappa^2}{n^2}+\frac{\xi^2(1-\xi)^2\theta\eta n^3\kappa^2d}{m\lambda_0} + \frac{\eta^2\theta^2 n \kappa ^2 \lambda_0 \xi^4}{m^{4}} +  \right. \\
    &\left. \quad\quad\quad\quad\quad\quad\quad \frac{\xi^2(1-\xi)^2\theta^2\eta^2 n^2  \kappa^2 d}{m^3\lambda_0 }+ \frac{\xi^2(1-\xi)^2\theta^2\eta^2 \kappa^2 \lambda_0}{m^3} + \frac{\xi^2(1-\xi)^2\theta^2\eta^2 n^2  \kappa^2 d}{m^2\lambda_0 }\right) \\
\end{align*}
With probability $1-\delta$, it holds for all $t\in [T]$ that
\begin{align*}
    \norm{\w_{r,t}-\w_{r,0}}_2 & \leq O\paren{\frac{t \lambda_0^2 K}{\delta n\sqrt{m}}} + \eta\theta  T\cdot O\left(\frac{\theta\eta\lambda_0^3 \xi^2 \kappa^2}{n^2}+\frac{\xi^2(1-\xi)^2\theta\eta n^3\kappa^2d}{m\lambda_0} + \frac{\eta^2\theta^2 n \kappa ^2 \lambda_0 \xi^4}{m^{4}} +  \right. \\
    &\left. \quad\quad\quad\quad\quad\quad\quad \frac{\xi^2(1-\xi)^2\theta^2\eta^2 n^2  \kappa^2 d}{m^3\lambda_0 }+ \frac{\xi^2(1-\xi)^2\theta^2\eta^2 \kappa^2 \lambda_0}{m^3} + \frac{\xi^2(1-\xi)^2\theta^2\eta^2 n^2  \kappa^2 d}{m^2\lambda_0 }\right) \\
\end{align*}
To enforce $\norm{\w_{r,t}-\w_{r,0}}_2 \leq R := O\paren{\frac{ \kappa\lambda_0}{n}}$, we then require
\begin{align*}
    m = \Omega\paren{\frac{n^3K^2}{\lambda_0^4\delta^2\kappa^2}\max\{n, d\}}
\end{align*}

More specific,
\begin{align*}
\norm{\w_{r,t}-\w_{r,0}}_2 \leq R_t := O\paren{\frac{t \kappa\lambda_0}{n}}     \\
\norm{\w_{r,t}-\w_{r,t-E}}_2 \leq R_E := O\paren{\frac{E \kappa\lambda_0}{n}}
\end{align*}

\clearpage
\end{document}

%% file: math_commands.tex
\usepackage{amsmath,amsfonts,bm}

\def\eqref#1{equation~\ref{#1}}

\def\1{\bm{1}}

\DeclareMathAlphabet{\mathsfit}{\encodingdefault}{\sfdefault}{m}{sl}
\SetMathAlphabet{\mathsfit}{bold}{\encodingdefault}{\sfdefault}{bx}{n}

\newcommand{\E}{\mathbb{E}}

\newcommand{\R}{\mathbb{R}}

\newcommand{\argmin}{\mathrm{argmin}}

\def\R{\mathbb{R}}

\newcommand{\norm}[1]{\left\|#1\right\|}

\newtheorem{thm}{Theorem}[section]
\newtheorem{lem}{Lemma}[section]

\newtheorem{asmp}{Assumption}[section]

\usepackage{amsmath,amsfonts,bm}

\newcommand{\x}{\mathbf{x}}
\newcommand{\y}{\mathbf{y}}

\newcommand{\X}{\mathbf{X}}

\newcommand{\W}{\mathbf{W}}
\newcommand{\w}{\mathbf{w}}
\newcommand{\U}{\mathbf{u}}
\newcommand{\h}{\mathbf{H}}
\newcommand{\xij}{\hat{\mathbf{x}}_i^{(j)}}
\newcommand{\paren}[1]{\left(#1\right)}
\newcommand{\indy}[1]{\mathbb{I}\left\{#1\right\}}

\newcommand{\inner}[2]{\left\langle#1, #2\right\rangle}

\def\eqref#1{equation~\ref{#1}}

\def\1{\bm{1}}

\DeclareMathAlphabet{\mathsfit}{\encodingdefault}{\sfdefault}{m}{sl}
\SetMathAlphabet{\mathsfit}{bold}{\encodingdefault}{\sfdefault}{bx}{n}